%% file: main.tex
\pdfoutput=1
\documentclass[lettersize,journal]{IEEEtran}
\usepackage{amsmath,amsfonts}
\usepackage{algorithmic}
\usepackage{algorithm}
\usepackage{array}
\usepackage[caption=false,font=normalsize,labelfont=sf,textfont=sf]{subfig}
\usepackage{textcomp}
\usepackage{stfloats}
\usepackage{url}
\usepackage{verbatim}
\usepackage{graphicx}
\usepackage{multicol}
\usepackage{multirow}
\usepackage[numbers,sort&compress]{natbib}
\usepackage{hyperref}  
\usepackage{cleveref}  
\usepackage{balance}   
\usepackage{booktabs}
\usepackage{dsfont}

\def\BibTeX{{\rm B\kern-.05em{\sc i\kern-.025em b}\kern-.08em
    T\kern-.1667em\lower.7ex\hbox{E}\kern-.125emX}}

\begin{document}

\title{\textit{Prior-SG}: Task and Prior Driven Region Segmentation\\for Scene Graphs in Arbitrarily-Structured Environments}
\author{Giorgio Tonetti, Laurent Kneip, Abel Gawel, and Marco Hutter%
\thanks{Giorgio Tonetti, Laurent Kneip, Abel Gawel, and Marco Hutter are with the RAI Institute. Additionally, Giorgio Tonetti and Marco Hutter are with ETH Zurich.}%
}
\markboth{Preprint under review}%
{Tonetti \MakeLowercase{\textit{et al.}}: Prior-SG}

\maketitle

\begin{abstract}

Hierarchical 3D scene graphs are a promising representation for high-level spatial reasoning in autonomous mobile platforms. However, existing extraction frameworks typically rely on purely local visual clustering or strict geometric heuristics, such as wall-separated rooms, which fail in open-plan or arbitrarily-structured environments. We propose \textit{Prior-SG}, a task- and prior-driven framework that casts scene graph generation fundamentally as a probabilistic alignment problem. As the robot explores, it continuously aggregates an incoming RGB-D sensor stream into a physically grounded \textit{Instance Graph} utilizing a multi-scale, open-vocabulary feature fusion strategy. The system then infers the high-level functional semantics of this map through a Maximum A Posteriori (MAP) estimate, guided by a \textit{Prior Graph}—a logical expectation of the environment's structure and task-relevant vocabulary synthesized dynamically by a Large Language Model. By optimizing a Markov Random Field that fuses heterogeneous experts (visual, geometric, and discrete objects) with these topological priors, the system resolves local perceptual ambiguities. We validate this approach across diverse simulated residential datasets and large, open-plan real-world environments. \textit{Prior-SG} achieves state-of-the-art semantic region segmentation accuracy compared to recent baselines, robustly delineates distant functional boundaries in the absence of physical walls, and uniquely provides zero-shot ontological flexibility, enabling the robot to entirely restructure its spatial partitioning based on a given high-level task.
\end{abstract}

\begin{IEEEkeywords}
Hierarchical Scene Graphs, Open-Vocabulary, Task-Driven, Large Language Models, Outdoor Navigation
\end{IEEEkeywords}

\input{sections/introduction}

\input{sections/related_works}

\input{sections/method}

\input{sections/results}
\input{sections/conclusion}


\balance 

\bibliographystyle{IEEEtranN}
\bibliography{main}



\end{document}

%% file: sections/introduction.tex
\section{Introduction}
\label{sec:introduction}

Human understanding of our surroundings is inherently hierarchical. We do not perceive the physical world as a collection of geometric primitives, but rather as a structured composition where distinct parts form objects, objects populate regions, and regions aggregate into complex environments. This mental abstraction allows us to reason about space efficiently, filtering out irrelevant details to focus on semantic relationships and context. To endow autonomous robotic systems with a similar level of spatial intelligence, we must move beyond flat semantic labels and embrace hierarchical data structures that mirror this cognitive organization. 

By bridging the gap between low-level geometry and high-level abstraction, hierarchical 3D scene graphs offer a compelling, flexible representation that allows us to address a broad spectrum of problems and tasks \cite{armeni20193d}. The graph structure innately enforces sparsity, which directly enables scalability to large-scale environments \cite{hughes2022hydra,hughes2024foundations} and computational efficiency in down-stream tasks. Simultaneously, its semantic interpretability enables direct serialization into natural language, facilitating seamless integration with Large Language Models (LLMs) for open-ended spatial reasoning \cite{chang2025ashita,werby2024hierarchical,gorlo2025describe} and environment interaction. Scene graphs transform complex geometry into a logical framework where agents can plan tasks and interactions with conceptual clarity \cite{chang2025ashita,werby2024hierarchical,ray2024task,rotondi2025fungraph,zhang2025open,maggio2024clio}.

Existing scene graph extraction approaches, however, often trade off generality for environment-specific performance. Especially for higher-level semantics, such as inferring regions, methods typically rely on strong, hard-coded geometric assumptions about the structure of their target environments (e.g., wall-separated rooms\cite{hughes2022hydra,werby2024hierarchical}, strict street topologies \cite{greve2024collaborative,deng2024opengraph}) and lack the ability to take into account higher-level topological priors about the environment. Because these methods depend on geometric heuristics, they fail in open-plan architecture or arbitrarily-structured spaces. Beyond this geometric rigidity, frameworks relying on purely bottom-up visual clustering suffer from sensory noise and semantic smearing \cite{maggio2024clio}. Furthermore, they are fundamentally limited by their reliance on dense, proximal observations, prohibiting efficient operation in large-scale environments where semantics even for distant, partially observed regions are required to inform navigation decisions.

To address these limitations, we introduce \textit{Prior-SG}, a framework designed to harmonize high-level semantic expectations with sensory observations. We are inspired by human spatial reasoning, which fundamentally operates differently than traditional mapping systems \cite{tversky1993cognitive}: we do not understand our surroundings purely through local geometry or isolated visual signals, but rather interpret them using a task-driven semantic prior of the expected environmental structure \cite{mandler2014stories}. To replicate this cognitive capability, we frame scene graph generation as a probabilistic alignment problem. The incoming sensor stream is continuously aggregated into an \textit{Instance Graph} that captures the local physical information of the observed environment. We then infer the task-relevant functional semantics of the environment through a Maximum A Posteriori (MAP) estimate. This inference is explicitly guided by a \textit{Prior Graph}—a probabilistic expectation synthesized by a Large Language Model that defines both the required semantic contents and the expected structural layout of the environment. For instance, when operating in a residential home, this prior establishes that a kitchen is likely present, that it structurally connects to a dining room, and that it typically contains objects like a refrigerator and a stove. To enable this alignment even when observations are sparse or distant, we employ a multi-scale feature fusion strategy that extracts language-aligned descriptors from the environment at varying levels of granularity. Crucially, this open-vocabulary architecture enables highly adaptable behavior. By generating the prior dynamically, we can fine-tune the exact set of relevant regions and objects to the specific mission the robot needs to solve, providing zero-shot ontological flexibility.

The specific contributions of this work are as follows:
\begin{itemize} 
    \item A formal definition of the scene graph generation task as a probabilistic alignment problem. We decouple the representation into a task-conditioned \textit{Prior Graph} generated by an LLM and a physically grounded \textit{Instance Graph} that aggregates local observations, inferring the high-level semantics via a MAP estimate (\Cref{sec:problem_formulation}).
    \item A multi-scale feature pyramid strategy that extracts scale-aware, open-vocabulary descriptors from 2D images. This overcomes the semantic smearing of global-only feature extraction and enables the robust prediction of distant, partially observed regions (\Cref{sec:instance_graph}). 
    \item A robust, global inference framework using graph cuts that fuses heterogeneous modalities such as visual textures, discrete objects, structural geometry, and soft topological regularizers to resolve local perceptual ambiguities and extract task-relevant regions in arbitrarily-structured environments (\Cref{sec:graph_cuts}). 
\end{itemize}

To validate these contributions, we comprehensively evaluate \textit{Prior-SG} (\Cref{sec:experiments}) across diverse simulated residential datasets and large-scale open-plan real-world environments, including a high-traffic train station. Our results demonstrate that the system achieves state-of-the-art segmentation accuracy, successfully infers functional boundaries in the absence of physical walls, and enables zero-shot, task-relevant spatial partitioning of physical environments given high-level natural language tasks.

%% file: sections/related_works.tex
\section{Related Works}
\label{sec:related_works}

The evolution of robotic spatial perception reflects a shift from purely metric maps to semantically augmented, hierarchical world representations. Within this context, scene graphs~\cite{bae2023survey,rotondi20263dscenegraphsopen} are often considered for their ability to compactly describe environments at varying levels of abstraction and using natural language, thereby forming a bridge towards navigation and reasoning~\cite{yin2024sgnav,kim20193,rana2023sayplan}, action planning~\cite{catalano2025survey}, or even human-robot interaction~\cite{ma2025llmsstep3dworld}. We group our prior work section into four parts: Foundations of 3D scene graphs; modern extensions including dynamics, affordance, or functional relationships; open-vocabulary adaptations via foundation models; and task-driven representations.

\subsection{Hierarchical 3D Scene Graphs for Robotic Perception}
Scene graphs originate in the computer vision community for structured image understanding \cite{johnson2015image, krishna2017visual} and were later extended to 3D robotics. \citet{armeni20193d} formalize 3D Scene Graphs (3DSG) as a four-layer hierarchy (building, room, object, camera), grounding semantics in geometry in an offline process.
%
%
\citet{rosinol2021kimera} and \citet{hughes2022hydra,hughes2024foundations} later introduce real-time 3D scene graph frameworks by separating efficiently discovered topological free space (``Places'') from semantic regions (``Rooms'') to support real-time navigation. \textit{Hydra} \cite{hughes2022hydra,hughes2024foundations} in particular enables online construction of Dynamic Scene Graphs (DSGs) using a sparsified Generalized Voronoi Diagram \cite{oleynikova2018sparse}. In a concurrent effort, \citet{bavle2022situational,bavle2023sgraphsplus,bavle2025sgraphs2} and \citet{millanromera2025metric} develop a line of work aiming for real-time situational graphs. They distinguish this representation from scene graphs by formulating it as a factor graph, allowing the metric and semantic states of the environment to be jointly optimized during online operation. These systems rely on closed vocabulary detectors and make strong assumptions about the geometric structure of the environment: wall-separated rooms extractable via geometric area segmentation approaches \cite{schwertfeger2019}. Large-scale outdoor alternatives exist, but assume lane-based topologies~\cite{deng2024opengraph,greve2024collaborative}, closed vocabulary-based terrain classification~\cite{samuelson2025towards,samuelson2026terra}, or detectable object-level hierarchies~\cite{nyffeler2025hierarchical}. More flexible alternatives include ontology-driven reasoning with Logic Tensor Networks \cite{strader2024indoor} and information-theoretic clustering in \textit{Clio} \cite{maggio2024clio}, which avoid fixed geometric assumptions. To the best of our knowledge, \textit{Clio}~\cite{maggio2024clio} is the most related to our method and serves as a baseline. Note that the present introduction is not exhaustive, and direct learning-based node and edge regression from point clouds~\cite{Wald_2020_CVPR} or incremental frame-wise fusion using Graph Neural Networks~\cite{Wu_2021_CVPR} have been proposed, too. However, these methods often focus on smaller-scale indoor environments, and do not offer the flexibility of task-driven region definition.

\subsection{Extensions}

Several extensions to the plain hierarchical scene graph model coined by~\citet{armeni20193d} exist. For example, ``Dynamic Scene Graphs'' aim at additionally modeling dynamic information of the object nodes in the graph~\cite{rosinol20203d,Schmid2024Khronos}. Further methods achieving life-long mapping and tracking long-term changes in the environment are introduced by \citet{gorlo2025describe}, \citet{looper2023vsg}, and \citet{behrens2025lost}. Another common extension consists of augmenting scene graphs by relational or functional attributes. For example, functional extensions such as \textit{FunGraph} \cite{rotondi2025fungraph} and \textit{OpenFunGraph} \cite{zhang2025open} infer affordances from language priors. \citet{kim20193} propose object-centric relational graphs to enable probabilistic reasoning, though without large-scale hierarchy. The scene graph prediction methods of \citet{Wald_2020_CVPR} and \citet{hou2025fross} augment hierarchical indoor representations by relational attributes. \citet{delitzas2024scenefun3d} propose a method for fine-grained functionality and affordance understanding in 3D scenes. While interesting and useful in terms of semantic expressivity, the above-mentioned approaches remain largely bottom-up in construction.

\subsection{Open-Vocabulary Semantic Mapping}

Modern semantic mapping leverages large vision-language models to enable open-vocabulary perception. Class-agnostic segmentation with SAM and FastSAM \cite{kirillov2023segment,zhao2023fast}, open-set detection through Grounding DINO and SAM3 \cite{liu2024grounding,carion2025sam3segmentconcepts}, and lightweight encoders such as  CLIP, MobileCLIP, and SigLIP \cite{radford2021learning,vasu2024mobileclip,zhai2023sigmoid,tschannen2025siglip} provide rich visual primitives. 
A common paradigm is ``feature painting'', lifting 2D embeddings into dense 3D representations. LERF \cite{lerf2023} optimizes a volumetric language field for dense querying, while explicit alternatives such as VLMaps \cite{huang23vlmaps}, OpenScene \cite{Peng_2023_CVPR}, and LSeg \cite{li2022languagedriven} project features directly onto voxel or point-based maps. Because storing high-dimensional vectors per voxel is highly memory-intensive, recent methods like FindAnything \cite{laina2025findanything} scale this approach by aggregating pixel-wise features via 2D segmentation before fusing them into memory-efficient, object-centric volumetric submaps.
Moving beyond raw volumetric maps, recent work integrates these semantics directly into discrete graph structures. \textit{ConceptGraphs} \cite{gu2024conceptgraphs} and \textit{Clio} \cite{maggio2024clio} fuse instance masks into open-vocabulary object graphs, and \textit{HOV-SG} \cite{werby2024hierarchical} and OVIGo-3DHSG \cite{linok25ov} additionally enable hierarchical querying. An in-depth analysis of feature-fusion and view selection techniques is provided by \citet{kassab2024bare}. \citet{Koch_2024_CVPR} combine image-based open-vocabulary feature extraction with point cloud-based scene graph prediction, again leveraging an LLM in order to add relational attributes to the graph. The method of \citet{deng2024opengraph} leverages open-vocabulary detectors for large-sale outdoor scene graphs. More recently, \citet{rotondi26keysg} augment scene graphs by a keyframe graph to efficiently leverage VLMs to extract scene information and enable task-agnostic reasoning and planning.

\subsection{Task-Driven and Probabilistic Graph Reasoning}

Task-dependent representations treat scene abstraction as inherently goal-conditioned. \textit{Clio} \cite{maggio2024clio} applies the Agglomerative Information Bottleneck \cite{tishby2000information,slonim1999agglomerative} to derive minimal task-sufficient graphs, while \textit{ASHiTA} \cite{chang2025ashita} uses an LLM to decompose tasks and request appropriate graph granularity. In parallel, LLMs have been studied as sources of structural priors. They encode object–room co-occurrence statistics \cite{chen2022extracting} and can be extended to explicit spatial reasoning as in \textit{SpatialVLM} \cite{Chen_2024_CVPR}, though benchmarks reveal limitations in generating spatially consistent graphs without structure \cite{yang-etal-2025-llm-meets}. LLMs have also been used for planning and probabilistic inference: \textit{SayPlan} \cite{rana2023sayplan} grounds long-horizon planning in scene graphs, while \citet{nafar2025extracting} and \citet{abbasi2024believe} demonstrate that LLMs can provide conditional probability estimates for Bayesian reasoning.

\subsection{Gap and Contribution}

Existing spatial perception systems typically fall into two disjoint paradigms: they either construct 3D graphs heavily reliant on rigid geometric heuristics and raw visual clustering \cite{hughes2022hydra,werby2024hierarchical} or extract abstract structured knowledge from language models for planning without dense physical grounding \cite{rana2023sayplan,nafar2025extracting}. Consequently, current frameworks lack an explicit mechanism to reconcile noisy, uncertain sensory observations with probabilistic structural expectations, and they fail to adapt their spatial partitioning to the specific task at hand.

We address this gap by proposing a task- and prior-driven approach to region segmentation capable of operating in arbitrarily-structured environments. Rather than treating scene understanding primarily as a reconstruction task bound by fixed spatial ontologies, we cast scene understanding fundamentally as a probabilistic alignment problem. Inspired by classical energy minimization methods such as graph cuts \cite{boykov2002fast}, our framework aligns the incrementally observed physical environment against a task-conditioned structural prior. This formulation enables local perceptual ambiguities to be resolved through global structural context, unifying open-vocabulary semantic perception and probabilistic reasoning within a single framework.

%% file: sections/method.tex
\section{The \textit{Prior-SG} Scheme}
\label{sec:problem_formulation}
\begin{figure*}[h!!]
    \centering
    \includegraphics[width=\textwidth]{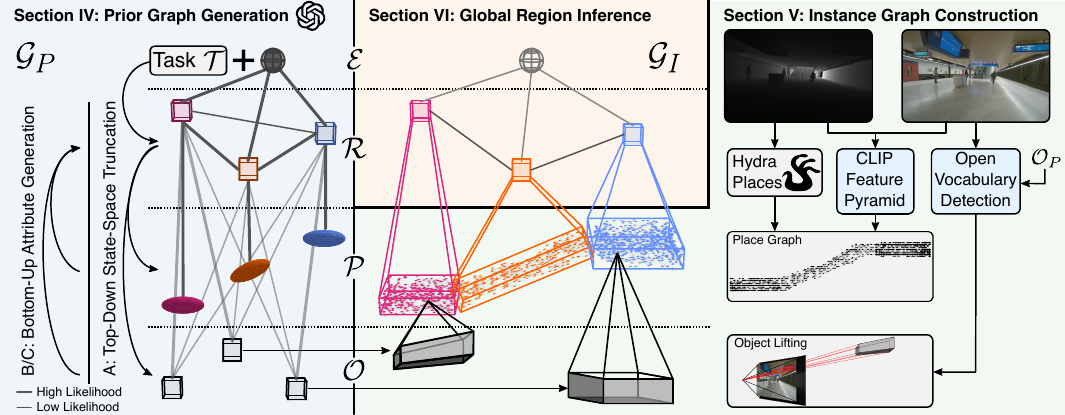}
\caption{System architecture of the \textit{Prior-SG} framework. As detailed in \Cref{sec:prior_graph}, the Prior Graph $\mathcal{G}_P$ (left) is synthesized by a Large Language Model conditioned on a high-level task and environment. This involves top-down state-space truncation (A) to define the expected hierarchical ontology, and bottom-up attribute generation (B/C) to assign intrinsic visual templates, geometric priors, and structural edges. Concurrently, as described in \Cref{sec:instance_graph}, raw RGB-D sensor streams are processed to incrementally build the physically observable layers of the Instance Graph $\mathcal{G}_I$ (right). Continuous free space is abstracted into a topological \emph{place} graph, while objects and multi-scale visual textures are extracted using open-vocabulary foundation models. Finally, \Cref{sec:graph_cuts} details the global region inference (center), where accumulated observations are probabilistically aligned against the prior expectation to extract high level semantics. A Markov Random Field optimization computes the optimal spatial partitioning field, explicitly instantiating the \emph{region} layer to complete the task-conditioned scene graph.}
    \label{fig:dual_graph_system_overview}
\end{figure*}

To realize this structural alignment, we draw inspiration from human spatial cognition. We do not understand our surroundings purely through local geometry or bottom-up visual signals. Arguably, we interpret them using a task-driven semantic prior of the expected environmental structure. To replicate this cognitive capability, we propose a prior-guided system. By applying a task-driven, open-vocabulary prior, the system resolves local perceptual ambiguities without relying on rigid geometric heuristics or hard-coded rules.

To formalize this approach, we define a spatial hierarchy $\mathcal{L} = (\mathcal{E}, \mathcal{R}, \mathcal{P}, \mathcal{O})$, which is vertically depicted along the central divider of \Cref{fig:dual_graph_system_overview}. The bottom layers, \emph{objects} $\mathcal{O}$ and \emph{places} $\mathcal{P}$, represent the physically observable elements of a scene. The structural layers, composed of \emph{regions} $\mathcal{R}$ within a root \emph{environment} $\mathcal{E}_0$, represent high-level semantics. They emerge directly from the semantic segmentation of the \emph{place} layer. However, for a specific instance of an environment, the space of valid functional breakdowns of the \emph{place} layer is effectively open-ended. Depending on the task, we may be interested in different interpretations of the environment. 
For example, consider an airport terminal (\emph{environment} $\mathcal{E}_0$) where physical seating benches and trash cans (\emph{objects} $\mathcal{O}$) are distributed across the navigable floor (\emph{places} $\mathcal{P}$). The optimal way to group these physical places into semantic zones (\emph{regions} $\mathcal{R}$) varies by application: an autonomous passenger transporter requires fine-grained partitions separating individual departure gates, whereas a cleaning robot operating in the exact same physical space requires a detailed segmentation of distinct dining and retail zones. We must therefore condition the segmentation on both the \emph{environment} $\mathcal{E}_0$ and the specific task $\mathcal{T}$ to be completed.

To operationalize this task-conditioned segmentation, we formulate this inference as finding the optimal partitioning field $L^*$—a discrete labeling function that assigns a region category to each node in the observed \emph{place} subgraph. This partition is visually depicted as the discrete coloring of the \emph{place} nodes in the Instance Graph of \Cref{fig:dual_graph_system_overview}. Formally, we seek the partition that maximizes the posterior probability given our sensor stream $\mathcal{Z}$, the global environment $\mathcal{E}_0$, and the task $\mathcal{T}$:
\begin{equation}
    \label{eq:map_problem}
    L^* = \arg \max_L P(L \mid \mathcal{Z}, \mathcal{E}_0, \mathcal{T})
\end{equation}
Since there is no well-defined probability measure to jointly model sensor streams ($\mathcal{Z}$), abstract linguistic concepts ($\mathcal{E}_0, \mathcal{T}$), and \emph{place} labels ($L$), we formulate a surrogate objective by projecting these heterogeneous variables into a unified probabilistic graphical model. 
In \Cref{sec:prior_graph}, we address the challenge of extracting the probability distribution over valid spatial configurations implicitly defined in $(\mathcal{E}_0, \mathcal{T})$. We define a generative mapping 
\begin{equation}
\label{eq:context_to_prior_graph_map}
    (\mathcal{E}_0, \mathcal{T}) \mapsto \mathcal{G}_P,
\end{equation}
employing a Large Language Model as an operator to translate the abstract context into a discrete, structured model. The resulting Prior Graph $\mathcal{G}_P$ acts as a discrete structural prior, explicitly encoding the expected region semantics, object containment, and topological structure. By substituting the context with its parametrized graph representation, we condition our inference on a tractable generative model:
\begin{equation}
    P(L \mid \mathcal{Z}, \mathcal{E}_0, \mathcal{T}) \approx P(L \mid \mathcal{Z}, \mathcal{G}_P).
\end{equation}
In \Cref{sec:instance_graph}, we introduce the Instance Graph $\mathcal{G}_I$ as the persistent representation of the observed environment. Specifically, we seek a perceptual aggregation mapping
\begin{equation}
    \mathcal{Z} \mapsto \mathcal{G}_I^{\text{obs}},
\end{equation}
where $\mathcal{G}_I^{\text{obs}}$ is the \textit{observable} subgraph of the Instance Graph, consisting of the \emph{object} and \emph{place} layers. During this construction process, the Prior Graph explicitly guides the extraction of open-vocabulary, language-aligned features from the sensor stream $\mathcal{Z}$. By assuming this observed subgraph encapsulates all relevant spatial and semantic evidence from the raw data, it serves as the discrete observation space for our graphical model, yielding the fully tractable posterior:
\begin{equation}
    P(L \mid \mathcal{Z}, \mathcal{E}_0, \mathcal{T}) \approx P(L \mid \mathcal{G}_I^{\text{obs}}, \mathcal{G}_P).
\end{equation}
In \Cref{sec:graph_cuts}, we apply Bayes' theorem to factorize this surrogate posterior into an observation likelihood and a structural prior:
\begin{equation}
    \label{eq:bayes_factorization}
    P(L \mid \mathcal{G}_I^{\text{obs}}, \mathcal{G}_P) \propto P(\mathcal{G}_I^{\text{obs}} \mid L, \mathcal{G}_P) P(L \mid \mathcal{G}_P).
\end{equation}
We cast the maximization of this factorized posterior as a Markov Random Field. By leveraging the language-aligned features and probabilistic structural rules established in the preceding mappings, we optimize the field $L$ using Graph Cuts over the \emph{place} subgraph, finally instantiating the \emph{region} layer $\mathcal{R}_I$ by using $L$ to induce the structural layer of the Instance Graph $\mathcal{G}_I^{\text{struct}}$.

\section{Prior Graph Extraction}
\label{sec:prior_graph}

As described in \Cref{sec:problem_formulation}, to solve the maximum a posteriori objective defined in \cref{eq:map_problem}, we require a mechanism to define the mapping in \cref{eq:context_to_prior_graph_map}. While the desired semantic categories depend on the task, the exact physical distribution of an environment's structure is effectively open-ended because every physical instance is unique. For example, while every airport terminal features a unique architectural layout, they all share common structural motifs—such as ticketing areas connecting to security checkpoints, which subsequently lead to departure gates. To approximate this unbounded structural space into a discrete, computationally tractable representation, we utilize a Large Language Model (LLM) as a common-sense approximator. It condenses the large structural variance into a discrete set of highly probable, discriminable rules. We formalize this expectation as the Prior Graph, defined as $\mathcal{G}_P = (V_P, E_P)$. 

\subsection{Top-Down State Space Truncation}
The construction of the node set $V_P$ proceeds top-down to bound the state space to a finite, task-relevant vocabulary. Conditioned on $\mathcal{E}_0$ and $\mathcal{T}$, we first prompt the LLM to enumerate the most probable functional zones, generating the prior \emph{region} set $\mathcal{R}_P \subset V_P$. This acts as a semantic relevance filter, clustering the open-ended semantic space into a focused target set. Next, for each instantiated region $r \in \mathcal{R}_P$, we generate the observable nodes, consisting of \emph{objects} $\mathcal{O}_P$ and \emph{places} $\mathcal{P}_P$. By establishing the region hypotheses first, we prompt the LLM to populate the prior \emph{object} set $\mathcal{O}_P \subset V_P$ as the union of two object subsets: objects that are strictly relevant to the task $\mathcal{T}$, and objects that provide high discriminative power for the region segmentation objective.
To anchor these abstract regions in physical space, we instantiate a single prototypical \emph{place} node $p_r \in \mathcal{P}_P$ for each region, establishing a practical unimodal spatial constraint per region concept.

\subsection{Intrinsic Node Attributes}
Having established the structural skeleton, we assign intrinsic attributes $\mathcal{D}(v_P)$ to each observable node $v_P \in \mathcal{O}_P \cup \mathcal{P}_P$ of the prior graph. These attributes serve as a reference dictionary to evaluate observation likelihoods. To enable probabilistic alignment, the abstract prior attributes and the physical sensor observations must be encoded into shared mathematical spaces. While this concept naturally extends to any sensory modality, our implementation specifically projects semantic appearance into a joint vision-language embedding space, and structural geometry into a set of mathematical parameters. To generate vision-aligned appearance descriptors $d_{\text{vis}}(v_P)$, we process the text labels of both \emph{object} and \emph{place} nodes through a CLIP text encoder. We utilize standard observation phrases (e.g., ``A picture of a [label]'') to average and stabilize the text embeddings, projecting the abstract linguistic concepts directly into the same embedding space utilized by the visual perception system. For \emph{place} nodes, we additionally generate geometric descriptors $d_{\text{geo}}(v_P)$. We prompt the LLM to define the expected local geometry, such as planarity or inclination with respect to gravity, and parse these responses into parametric mathematical parameters (e.g., expected angles and variances). Unlike existing baselines that impose hard structural boundaries through geometric heuristics, our geometric priors act as statistical properties that provide soft evidence for or against specific region hypotheses.

\subsection{Probabilistic Structural Edges}
Beyond the physical appearance of individual nodes, the MAP formulation requires mathematical rules governing how these elements compose and connect. We define the structural descriptors of the prior graph as its edge set $E_P$, which explicitly encodes these relational probability distributions to serve as penalty terms in the graph cut solver. First, we define Containment Priors via hierarchical edges. We query the LLM to estimate the conditional likelihood matrix $P(o \mid r,\mathcal{E}_0)$, representing the probability of an object $o$ existing within a region $r$. We employ a soft probabilistic matrix rather than binary associations because certain objects may characterize multiple regions or be task-relevant but non-discriminative. This yields a probabilistically expressive signature for each region. We simultaneously estimate the global region existence prior $P(r \mid \mathcal{E}_0)$. Second, we define Topological Priors via adjacency edges by estimating the transition likelihood matrix $P(r_i \leftrightarrow r_j \mid \mathcal{E}_0)$. This matrix encodes the logically plausible spatial layout, resolving local ambiguities during inference when visual cues alone indicate a departure from one region but cannot distinguish between candidate destination regions.

At the conclusion of this generative phase, the system holds a formally defined Prior Graph $\mathcal{G}_P$. This graph acts as a structured probability distribution populated with intrinsic observation templates and relational structural edges, providing the necessary foundation to guide the bottom-up feature extraction detailed in \Cref{sec:instance_graph} and to regularize the global alignment in \Cref{sec:graph_cuts}.

\section{Constructing the Instance Graph}
\label{sec:instance_graph}

To solve the maximum a posteriori objective, we must aggregate the raw RGB-D sensor stream $\mathcal{Z}$ into a persistent state representation that can be directly evaluated against the expectations established by $\mathcal{G}_P$. To this end, we incrementally construct an Instance Graph $\mathcal{G}_I$, explicitly guiding the construction of its observable layers—composed of \emph{objects} $\mathcal{O}_I$ and \emph{places} $\mathcal{P}_I$—using the expected semantics encoded in the Prior Graph. The primary goal of this construction is to gather task-relevant sensory information and aggregate it into observation descriptors (e.g., $\hat{d}_{\text{vis}}, \hat{d}_{\text{geo}}$). To infer the regions in a later stage, these descriptors must be explicitly encoded into the same compatible latent and parametric spaces as the intrinsic attributes of the Prior Graph's nodes.

\subsection{Instantiating the Physical Backbone}
We require a stable 3D topological backbone to host our features. To capture both discrete objects and continuous navigable space, we construct this backbone using two parallel representations.
First, to instantiate objects ($\mathcal{O}_I$), instead of unconstrained open-vocabulary discovery, we constrain an open-set object detector \cite{liu2024grounding} by prompting it explicitly with the prior object set $\mathcal{O}_P$ to avoid injecting irrelevant information into the graph. This guarantees that all instantiated 3D objects are task-relevant or discriminative. Detections are lifted to 3D and associated temporally via geometric intersection. 
Second, for place instantiation ($\mathcal{P}_I$), we represent continuous free space using established spatial mapping techniques. We adopt the \emph{place} extraction strategy introduced by Hydra \cite{hughes2022hydra,hughes2024foundations}, which incrementally constructs a topological graph of unoccupied space. Each \emph{place} node $p \in \mathcal{P}_I$ represents a spatial volume anchored at a 3D centroid $\mathbf{x}_p \in \mathbb{R}^3$.

\subsection{Multi-Scale Feature Pyramid}
Extracting language-aligned appearance descriptors for discrete objects is achieved efficiently by encoding their 2D bounding boxes. In contrast, characterizing the visual texture of unoccupied 3D space presents an inherent challenge. The semantic essence of a spatial volume is strongly view-dependent and lacks obvious image-space boundaries that could be leveraged to constrain the 2D semantic search space. To resolve this, we model each place as a 3D Gaussian volume, anchored at a centroid $\mathbf{x}_p$ and parameterized by an isotropic scalar covariance $r_Q$. Crucially, this radius $r_Q$ is an intrinsic property derived directly from the Prior Graph place node $p_r \in \mathcal{P}_P$, explicitly defining the expected physical size of the relevant semantic context for a given environment.

Both our target observations and our discrete image crops can be formally defined as states within a 3-dimensional parameter space. We define this state as a tuple $s = (\mathbf{u}, \sigma) \in \mathbb{R}^2 \times \mathbb{R}^+$, encapsulating a 2D pixel coordinate and a scale. For a given camera observation, projecting a place's 3D Gaussian into the image plane yields the target query state $s_p$ for that observation, defined as
\begin{equation}
s_p = (\mathbf{u}_p, \sigma_p) = \left(\pi(\mathbf{x}_p), \frac{f \cdot r_Q}{z_p}\right),
\label{eq:query_state}
\end{equation}
where $\pi(\cdot)$ denotes the perspective projection, $f$ is the camera focal length, and $z_p$ is the depth of the centroid relative to the camera. The theoretically optimal visual feature for this observation would be obtained by taking a square image crop centered exactly at $\mathbf{u}_p$ with a side length of $2\sigma_p$ and passing it through a vision-language encoder. However, dynamically computing this optimal feature on-the-fly is computationally intractable. A single frame may observe an arbitrary number of places, continuous SLAM pose refinements alter relative geometries (requiring expensive historical re-projections), and spatially adjacent places would yield highly redundant crops.

To approximate this optimal continuous mapping efficiently, we introduce a static Feature Pyramid approach, inspired by LERF \cite{lerf2023}. Conceptually, this pyramid is a hierarchical stack of image feature grids. Each ``layer'' in the stack evaluates the image using uniformly sized, overlapping crops, with the crop size increasing at each successive layer. This allows the system to capture both fine-grained local details at the bottom of the pyramid and broad regional context at the top. Instead of dynamic cropping, we extract a discrete pyramid of features upfront. For an input image $I \in \mathbb{R}^{H \times W \times 3}$, we construct a pyramid composed of discrete layers $l \in \{0, 1, \dots, L\}$. Each layer utilizes a sliding window of dimension $S_l \times S_l$, where $S_l$ increases with $l$ up to $\min(H,W)$. This defines an overlapping discrete grid of candidate crops with spatial indices $i \in \mathcal{I}_l$. The state of the $i$-th crop in the $l$-th layer is parameterized by a fixed pixel center and scale, $s_{i,l} = (\mathbf{u}_{i,l}, \sigma_l)$, where $\sigma_l = S_l / 2$.

Before performing any computationally expensive feature extraction, we ensure that these arbitrary grid crops actually represent meaningful volumes of physical space. Relying on the registered depth map, we extract the median depth $z_i$ within the spatial bounds of each candidate crop. Because the crop's side-length is intrinsically defined by its scale state as $2\sigma_l$, we apply standard pinhole camera geometry to estimate the physical 3D area $A_{\text{3D}}$ associated with the crop state $s_{i,l}$:
\begin{equation}
    A_{\text{3D}}(s_{i,l}) \approx \left(\frac{2\sigma_l \cdot z_i}{f}\right)^2.
    \label{eq:physical_area}
\end{equation}
We aggressively discard any crops where $A_{\text{3D}} < A_{min}$, effectively filtering out regions that only observe minute physical fragments (e.g., a tiny patch of a nearby wall). The surviving crops are resized and batch-processed through the CLIP encoder to yield latent descriptors $\mathbf{e}_{i,l} \in \mathbb{R}^D$. The completed Feature Pyramid acts as a static repository for the image's language-aligned features, effectively storing the encoded texture of the scene as a set of state-feature pairs:
\begin{equation}
    \mathcal{F} = \Big\{ (s_{i,l}, \mathbf{e}_{i,l}) \Big\}_{l=0,\dots, L}^{i \in \mathcal{I}_l}.
    \label{eq:feature_pyramid}
\end{equation}

\subsection{Semantic Integration Kernel}
With the feature pyramid established, we formulate an integration kernel to approximate the optimal feature for the place query $s_p$ by mapping it against the discrete database $\mathcal{F}$. A query experiences two primary types of mismatch against a candidate crop: spatial misalignment and scale deviation.

First, we compute the spatial relevance of a crop to the projected place. We model this spatial affinity $w_{\text{spa}}$ as a 2D Gaussian response over the image plane, evaluated between the query state and the crop state:
\begin{equation}
    w_{\text{spa}}(s_p, s_{i,l}) = \exp\left( - \frac{\| \mathbf{u}_p - \mathbf{u}_{i,l} \|_2^2}{2(\sigma_p^2 + \sigma_l^2)} \right).
    \label{eq:spatial_affinity}
\end{equation}
We explicitly do not normalize this distribution. Normalization would artificially inflate the weight of distant, irrelevant crops when no geometrically aligned crop exists within a specific layer.

Second, we penalize pyramid layers where the candidate crop scale $\sigma_l$ deviates from the prior-expected scale $\sigma_p$. We define a scale trust factor $w_{\text{sca}}$ that acts as a band-pass filter across the pyramid layers, modulated by a bandwidth parameter $\gamma$:
\begin{equation}
    w_{\text{sca}}(\sigma_p, \sigma_l) = \exp\left( - \frac{(\log \sigma_p - \log \sigma_l)^2}{\gamma} \right).
    \label{eq:scale_trust}
\end{equation}

To compute the final descriptor approximation, we define the effective integration weight $W(s_p, s_{i,l})$ for each crop as the direct product of its spatial affinity (\cref{eq:spatial_affinity}) and its scale trust (\cref{eq:scale_trust}). We aggregate the discrete latent embeddings $\mathbf{e}_{i,l}$ using these effective weights to produce $\mathbf{e}_p$, representing the isolated visual observation of the place from this specific frame. Concurrently, we evaluate the overall quality of this scale-matched observation by extracting a scalar trust value $T(p)$, computed as the sum of peak spatial affinities weighted by the layer trust:
\begin{equation}
    \mathbf{e}_p = \sum_{l=0}^{L} \sum_{i \in \mathcal{I}_l} W(s_p, s_{i,l}) \, \mathbf{e}_{i,l},
    \label{eq:single_observation_feature}
\end{equation}
\begin{equation}
    T(p) = \sum_{l=0}^{L} w_{\text{sca}}(\sigma_p, \sigma_l) \max_{i \in \mathcal{I}_l} \Big\{ w_{\text{spa}}(s_p, s_{i,l}) \Big\}.
    \label{eq:observation_trust}
\end{equation}
This operation aggregates the dense, multi-scale image pyramid $\mathcal{F}$ into a single descriptive vector $\mathbf{e}_p$ and a corresponding confidence score $T(p)$. These values encapsulate the single-frame observation and are subsequently passed to the temporal fusion module to iteratively update the persistent \emph{place} node descriptor $\hat{d}_{\text{vis}}$.

\subsection{Temporal Aggregation and Geometric Descriptors}
Over the course of a trajectory, nodes receive a sequence of observations over time. For a given \emph{place} $p$, the integration kernel sequentially outputs single-frame feature vectors and their associated confidence scores, yielding a historical set $\mathcal{H}_p = \{(\mathbf{e}_p^{(1)}, T^{(1)}), \dots, (\mathbf{e}_p^{(K)}, T^{(K)})\}$. To aggregate these measurements into a single persistent feature for each entity, we apply weighted-average temporal fusion strategies. For \emph{places}, we compute a trust-weighted average of historical observations to strictly favor high-quality scale matches, producing the persistent vision-aligned appearance descriptor:
\begin{equation}
    \hat{d}_{\text{vis}}(p) = \frac{\sum_{k=1}^K T^{(k)} \mathbf{e}_p^{(k)}}{\left\| \sum_{k=1}^K T^{(k)} \mathbf{e}_p^{(k)} \right\|_2}.
    \label{eq:temporal_fusion_places}
\end{equation}

For \emph{objects} $\mathcal{O}_I$, which rely on deterministic bounding boxes rather than probabilistic scale projections, observations are assigned a uniform trust. Instead of a trust-weighted average, we execute a centrality-based average over the historical set of object encodings $\mathbf{e}_o^{(k)}$. This isolates the most representative visual prototype and inherently filters out transient occlusions or misdetections:
\begin{equation}
    \hat{d}_{\text{vis}}(o) = \frac{\sum_{k=1}^K w_{\text{cent}}^{(k)} \mathbf{e}_o^{(k)}}{\left\| \sum_{k=1}^K w_{\text{cent}}^{(k)} \mathbf{e}_o^{(k)} \right\|_2},
    \label{eq:temporal_fusion_objects}
\end{equation}
where $w_{\text{cent}}^{(k)}$ represents the spatial centrality weight of the $k$-th observation within the feature cluster.

Finally, to evaluate the soft geometric priors $d_{\text{geo}}$ generated by the LLM, we must extract matching geometric descriptors $\hat{d}_{\text{geo}}(p)$ from the observations. Rather than relying on computationally expensive dense point clouds, we analyze the structural layout of the topological \emph{place} graph itself. For a given \emph{place} $p$, we define its topological neighborhood $\mathcal{N}(p)$ using standard graph-theoretic traversal, identifying the subset of the \emph{place} graph reachable within a maximum shortest-path distance threshold $\delta_{\text{max}}$. Formally:
\begin{equation}
    \mathcal{N}(p) = \{q \in \mathcal{P}_I \mid \text{dist}_{\mathcal{G}}(\mathbf{x}_p, \mathbf{x}_q) \le \delta_{\text{max}}\}
    \label{eq:neighborhood_support}
\end{equation}
where $\text{dist}_{\mathcal{G}}(\mathbf{x}_p, \mathbf{x}_q)$ denotes the shortest-path distance between the spatial centroids along the edge set $E_I^{\mathcal{P}}$. To extract the local structural properties, we compute the sample covariance matrix of the 3D centroids within this neighborhood. An eigendecomposition of this matrix directly yields the parametric measures, such as the local inclination angle $\theta_p$ relative to gravity, required to construct $\hat{d}_{\text{geo}}(p)$, which can then be evaluated against the geometric templates established in $\mathcal{G}_P$.

\section{Global Inference via Graph Cuts}
\label{sec:graph_cuts}

The preceding phases established a generative, top-down probabilistic rulebook ($\mathcal{G}_P$) and extracted a bottom-up spatial backbone annotated with aligned descriptors. The final objective is to instantiate the \emph{region} set $\mathcal{R}_I$ by finding the optimal labeling field $L: \mathcal{P}_I \rightarrow \mathcal{R}_P$ that maps the \emph{place} graph into discrete functional zones. 

As established in \cref{eq:bayes_factorization}, we formulate this as a Maximum A Posteriori (MAP) inference problem. By taking the negative logarithm of the factorized posterior, the product of probabilities transforms into a sum of energy potentials, mapping the inference directly to a Markov Random Field (MRF) energy minimization framework:
\begin{equation}
    E(L) = \sum_{p \in \mathcal{P}_I} \psi_U(p, L(p)) + \sum_{(u,v) \in E_I^{\mathcal{P}}} \psi_P(L(u), L(v)).
    \label{eq:mrf_energy}
\end{equation}
The unary potential $\psi_U$ measures the cost of assigning \emph{place} $p$ to \emph{region} $r \in \mathcal{R}_P$ based on local observations, while the pairwise potential $\psi_P$ acts as a topological regularizer over the set of spatial edges $E_I^{\mathcal{P}}$ connecting adjacent \emph{places} $(u,v)$.

\subsection{Unary Potentials via a Product of Experts}
Our observations consist of highly heterogeneous modalities: visual textures, local geometries, discrete object detections, and global context. Furthermore, physical spatial data is inherently noisy and incomplete; a \emph{place} might exhibit highly discriminative visual texture but lack any discernible objects. To fuse these disparate signals gracefully without forcing decisions from missing data, we model the joint observation likelihood using a Confidence-Weighted Log-Linear Pool (a Product of Experts). 

Let $\mathcal{K} = \{\text{vis}, \text{geo}, \text{obj}, \text{env}\}$ be the set of available expert modalities. We define the fused probability of assigning region $r$ to place $p$ as:
\begin{equation}
    P(r \mid p) \propto \prod_{k \in \mathcal{K}} P_k(r \mid p)^{\alpha_k(p)}.
    \label{eq:product_of_experts}
\end{equation}
The effective weight for each expert is $\alpha_k(p) = c_k(p) \cdot \lambda_k$, where $\lambda_k$ is a static global hyperparameter governing the modality's baseline reliability, and $c_k(p) \in [0, 1]$ is a dynamic, instance-specific observation confidence. Taking the negative logarithm of \cref{eq:product_of_experts} yields the additive unary potential:
\begin{equation}
    \psi_U(p, r) = - \sum_{k \in \mathcal{K}} \alpha_k(p) \log P_k(r \mid p).
    \label{eq:unary_potential}
\end{equation}
Crucially, if an expert lacks sufficient data for a specific place, its dynamic confidence drops to zero ($c_k(p) = 0$). This effectively drives the exponent in the product to zero (or the multiplier in the summation to zero), smoothly ignoring the expert without corrupting the joint distribution.

We formally define the derivations for the probability distributions $P_k$ and the confidence weights $c_k$ for the independent experts as follows:

\subsubsection{The Visual Expert ($P_{\text{vis}}$) and Open-Set Ambiguity}
The visual expert computes the cosine similarity 
\begin{equation}
    s_r = \hat{d}_{\text{vis}}(p)^\top d_{\text{vis}}(r)
\end{equation}
between the place's observed descriptor and the reference descriptors of the prior region concepts. However, real-world environments contain open-set clutter that the LLM did not explicitly hypothesize. To prevent the system from forcibly aligning irrelevant clutter to a known prior concept, we introduce a ``Null'' score $s_{\emptyset}$ (acting as a background noise floor) and apply a temperature-scaled Softmax over both the regions and the null class:
\begin{equation}
    p_r = \frac{\exp(\tau s_r)}{\exp(\tau s_{\emptyset}) + \sum_{j=1}^{|\mathcal{R}_P|} \exp(\tau s_j)}.
\end{equation}
The dynamic visual confidence $c_{\text{vis}}(p)$ is defined as the total probability mass successfully assigned to the valid prior concepts, $c_{\text{vis}}(p) = \sum_{j=1}^{|\mathcal{R}_P|} p_j$. If the observation is predominantly out-of-distribution, the mass flows to the null sink, and $c_{\text{vis}}(p)$ correctly diminishes. The final visual distribution is then obtained by normalizing out the null class:
\begin{equation}
    P_{\text{vis}}(r \mid p) = \frac{p_r}{c_{\text{vis}}(p)}.
\end{equation}

\subsubsection{The Geometric Expert ($P_{\text{geo}}$)}
While visual textures and discrete objects resolve many semantic ambiguities, certain functional zones (e.g., ramps, staircases) are fundamentally defined by their physical layout rather than their appearance. To integrate this structural evidence, the geometric expert evaluates the observed geometric descriptor $\hat{d}_{\text{geo}}(p)$ (e.g., the extracted local inclination angle $\theta_p$) against the LLM's parametric templates, which provide an expected mean $\mu_r$ and variance $\sigma_r^2$ for a given region $r$. We model this angular alignment using a von Mises distribution, where the concentration parameter $\kappa_r$ is inversely proportional to the prior variance:
\begin{equation}
    P_{\text{geo}}(r \mid p) \propto \exp \Big( \kappa_r \cos(\theta_p - \mu_r) \Big).
\end{equation}
Furthermore, the extracted angle $\theta_p$ is only meaningful if the underlying spatial neighborhood actually forms a coherent structural surface. To prevent unstructured noise from corrupting the inference, the geometric confidence $c_{\text{geo}}(p)$ reflects the reliability of the underlying covariance extraction. It is computed as the product of a support saturation function over the neighborhood $\mathcal{N}(p)$ defined in \cref{eq:neighborhood_support}, and the local shape anisotropy:
\begin{equation}
    c_{\text{geo}}(p) = \frac{|\mathcal{N}(p)|}{|\mathcal{N}(p)|+N_{\text{sat}}}\cdot\left( 1 - \frac{\lambda_3}{\lambda_1} \right)
\end{equation}
where $N_{\text{sat}}$ acts as a saturation constant reflecting the number of graph nodes required to form a stable 3D plane, and the ratio of the smallest to largest eigenvalues ($\lambda_3 / \lambda_1$) penalizes spherical, non-planar noise. This explicit formulation ensures that the geometric expert is smoothly ignored in unstructured or sparsely observed areas without corrupting the joint distribution.

\subsubsection{The Object Expert ($P_{\text{obj}}$) and Spatial Proximity Transfer}
To evaluate a region hypothesis using object detections, we must first ground the observed objects to the LLM's vocabulary. For an observed object $o \in \mathcal{O}_I$, we compute the cosine similarity between its visual descriptor $\hat{d}_{\text{vis}}(o)$ and the expected descriptor $d_{\text{vis}}(c)$ for all prior concepts $c \in \mathcal{O}_P$. Because the object layer $\mathcal{O}_I$ is instantiated using a prior-prompted open-vocabulary detector, we assume all instantiated objects map to a valid prior concept, circumventing the open-set ambiguity present in the continuous visual texture expert. We directly apply a temperature-scaled Softmax over the cosine similarities to yield a valid categorical distribution $P(c \mid o)$. Using the LLM's Containment Prior, we compute the object-to-region marginal probability: 
\begin{equation}
    P(r \mid o) = \sum_{c \in \mathcal{O}_P} P(r \mid c) P(c \mid o).
\end{equation}
To transfer this evidence from the discrete \emph{object} layer to the \emph{place} layer, we formally define a bipartite \emph{place}-\emph{object} proximity subgraph. A place $p$ is connected to an object $o$ only if their Euclidean spatial distance $\| \mathbf{x}_p - \mathbf{x}_o \|_2$ falls within a hard cutoff threshold $d_{\text{max}}$, defining the subset of anchored objects $\mathcal{O}_{\text{p}}$. Within this subgraph, the spatial relevance $w(p, o) \in [0, 1]$ decays organically with distance:
\begin{flalign}
    & w(p, o) = 
    \begin{cases} 
      \exp \left( - \frac{\| \mathbf{x}_p - \mathbf{x}_o \|_2}{\gamma_{\text{obj}}} \right), & \text{if } \rlap{$\| \mathbf{x}_p - \mathbf{x}_o \|_2 \le d_{\text{max}}$} \\
      0, & \text{otherwise.}
    \end{cases} &
\end{flalign}

To aggregate the localized evidence, we execute a proximity-discounted max-pooling operation over the anchored objects. For each region hypothesis $r$, we evaluate the distance-weighted evidence score across all nearby objects to yield an unnormalized region distribution $\tilde{P}_{\text{obj}}$:
\begin{equation}
    \tilde{P}_{\text{obj}}(r \mid p) = \max_{o \in \mathcal{O}_{\text{p}}} \Big( w(p, o) \cdot P(r \mid o) \Big).
\end{equation}
By embedding the spatial weight directly into the max-pool, the physical proximity of different objects organically shapes the unnormalized distribution. A closer, generic object will inherently dampen the relative score of a distant, definitive object. 

Mirroring the logic of the Visual Expert, the dynamic confidence $c_{\text{obj}}(p)$ is defined directly as the total volume of valid, distance-discounted probability mass in the proximity of the place:
\begin{equation}
    c_{\text{obj}}(p) = \sum_{j \in \mathcal{R}_P} \tilde{P}_{\text{obj}}(j \mid p).
\end{equation}
The final probability distribution $P_{\text{obj}}(r \mid p)$ for the Product of Experts is subsequently obtained by dividing the unnormalized scores by this confidence:
\begin{equation}
    P_{\text{obj}}(r \mid p) = \frac{\tilde{P}_{\text{obj}}(r \mid p)}{c_{\text{obj}}(p)}.
\end{equation}
This formulation yields a highly cohesive fusion framework: the normalization isolates the relative semantic distribution, while the confidence multiplier dynamically scales the expert's absolute authority in the global MRF based on spatial proximity. Most importantly, if a place exceeds $d_{\text{max}}$ to any instantiated object ($\mathcal{O}_{\text{p}} = \emptyset$), the unnormalized mass is zero, and the confidence strictly evaluates to $c_{\text{obj}}(p) = 0$, seamlessly removing the object expert from \cref{eq:unary_potential}.
\subsubsection{The Environment Prior Expert ($P_{\text{env}}$)}
Finally, the environment expert injects the global structural prior $P(r \mid \mathcal{E}_0)$ generated directly by the LLM. This provides a baseline spatial regularizer, gently favoring regions that are naturally expected to dominate the specified environment.

\subsection{Topological Regularization ($\psi_P$) and Graph Cuts}
To enforce spatial consistency, we define the pairwise potential $\psi_P$ as a Potts-like model. For any two adjacent places $(u,v) \in E_I^{\mathcal{P}}$, we penalize the transition if they are assigned to different regions ($L(u) \neq L(v)$). We modulate this penalty by a distance-based spatial decay weight $w_{\text{dist}}(u,v)$ and the topological transition log-likelihood extracted from the Prior Graph:
\begin{equation}
    \psi_P(L(u), L(v)) = 
    \begin{cases} 
      -\beta \cdot w_{\text{dist}}(u,v) \log P_{u \leftrightarrow v} & L(u) \neq L(v) \\
      0 & L(u) = L(v) 
    \end{cases}
    \label{eq:pairwise_potential}
\end{equation}
where $P_{u \leftrightarrow v} = P(L(u) \leftrightarrow L(v) \mid \mathcal{E}_0)$ is the transition likelihood matrix defined in $\mathcal{G}_P$.

Minimizing this MRF energy field efficiently typically relies on the $\alpha$-expansion algorithm. However, $\alpha$-expansion mathematically requires a metric cost. The complex topological rules generated by human common-sense inherently violate this requirement. For example, direct transitions between a Kitchen and a Corridor, and between a Corridor and a Bathroom, both incur low penalties. Yet, a direct transition between a Kitchen and a Bathroom incurs a high penalty, violating the triangle inequality mathematically. To strictly enforce these non-metric LLM topologies, we minimize the energy field using the $\alpha$-$\beta$-swap algorithm \cite{boykov2002fast}, which only evaluates binary label pairs and therefore avoids metric assumptions.

\subsection{Instantiating the Instance Layers}
Upon convergence, the optimization yields the MAP partitioning field $L^*$, assigning a discrete region label to every \emph{place} node. However, a single semantic label (e.g., ``Corridor'') may describe multiple physically disconnected instances within a large environment. 

To formally instantiate the discrete \emph{region} nodes of the Instance Graph, we isolate the vertex-induced subgraph $\mathcal{G}_I[\mathcal{P}^{(r)}]$ for each region label $r \in \mathcal{R}_P$, where $\mathcal{P}^{(r)} = \{p \in \mathcal{P}_I \mid L^*(p) = r\}$. We extract the Connected Components ($CC$) from these labeled subgraphs:
\begin{equation}
    \mathcal{R}_I = \bigcup_{r \in \mathcal{R}_P} CC \left( \mathcal{G}_I[\mathcal{P}^{(r)}] \right).
    \label{eq:connected_components}
\end{equation}
Each distinct physical component becomes a unique \emph{region} node in $\mathcal{R}_I$. Finally, hierarchical edges are established connecting each \emph{region} node down to its constituent \emph{places} and \emph{objects}, yielding a comprehensive, hierarchically segmented scene representation.

%% file: sections/results.tex
\section{Experimental Evaluation}
\label{sec:experiments}

In this section, we evaluate the \textit{Prior-SG} framework. Our primary objective is to validate its ability to construct hierarchical scene graphs by accurately extracting a \emph{region} layer from a physical \emph{place} layer in arbitrarily structured environments. 

Real-world spaces range from constrained residential layouts to large, open-plan public infrastructure. Traditional scene graph paradigms frequently fail at the extremes of this spectrum: geometric heuristics break down in the absence of physical dividing walls, and bottom-up visual clustering is highly susceptible to sensory noise, visual aliasing, and semantic smearing. We hypothesize that formulating region segmentation as a probabilistic inference problem, which aligns heterogeneous observations against a language- and task-driven topological prior, effectively resolves these perceptual ambiguities. This formulation aims to provide consistent semantic and spatial segmentation regardless of the underlying architectural layout.

To validate this hypothesis, our evaluation investigates four primary claims:
\begin{enumerate}
    \item \textbf{State-of-the-Art Robustness:} The prior-guided inference framework outperforms both geometric and information-theoretic open-vocabulary baselines in spatial and semantic metrics across diverse environmental scales.
    \item \textbf{Expert Synergy:} Segmentation can be significantly robustified by jointly fusing various heterogeneous signal sources, such as visual, geometric, topological, and spatial experts, to correctly resolve conflicting local signals.
    \item \textbf{Predictive Incremental Construction:} In large, wall-less real-world environments, the combination of multi-scale visual extraction and geometric constraints enables the online system to accurately predict distant regions and maintain crisp boundaries.
    \item \textbf{Task-Conditioned Flexibility:} Because the prior graph is generated dynamically via a Large Language Model, the framework exhibits zero-shot ontological flexibility, enabling it to entirely restructure the spatial partitioning of an identical physical environment based on high-level robotic personas.
\end{enumerate}

The remainder of this section is structured as follows. We first detail the experimental datasets, implementation parameters, baselines, and evaluation metrics (\Cref{subsec:datasets,subsec:implementation,subsec:baselines,subsec:metrics}). We then evaluate our claims by comparing system performance against the state-of-the-art (\Cref{subsec:sota_comparison}), ablating the constituent experts (\Cref{subsec:ablation_experts}), demonstrating open-plan robustness via real-world incremental scene graph construction (\Cref{subsec:incremental_construction}), highlighting task-conditioned spatial flexibility (\Cref{subsec:task_condition}), and evaluating the computational trade-offs of the feature extraction pipeline (\Cref{subsec:system_sensitivity}).

\subsection{Experimental Domains and Datasets}
\label{subsec:datasets}

To evaluate the region segmentation across a range of architectural layouts, we utilize four datasets spanning simulated and real-world environments. These datasets are selected to encompass constrained residential layouts, diverse semantic domains, and large-scale open-plan architecture.

\subsubsection{Habitat-Matterport 3D (HM3D)}
To establish baseline quantitative metrics against 3D ground truth in standard, room-based indoor spaces, we utilize the Habitat simulator \cite{szot2021habitat} with the Habitat-Matterport 3D (HM3D) dataset \cite{ramakrishnan2021hm3d}. This dataset provides realistic simulations of multi-floor residential apartments. To ensure a comprehensive evaluation, we define two distinct sets of trajectories for this environment, which we treat as separate datasets during our baseline comparisons. First, to allow for direct methodological benchmarking against prior art, we evaluate 8 \textit{Scan Trajectories} utilized by \citet{werby2024hierarchical}, which consist of exhaustive, highly detailed camera movements designed to scan the entire scene. Second, to evaluate real-world exploration dynamics, we recorded 8 custom \textit{Navigation Trajectories} that emulate the natural traversal of a home (e.g., \textit{Entrance} $\to$ \textit{Kitchen} $\to$ \textit{Living Room}). These aim to approximate the motion of an agent actively navigating a residential environment for the first time.

\subsubsection{TartanGround}
To evaluate the system's adaptability across a high diversity of semantic environments outside of standard residential layouts, we incorporate the TartanGround dataset \cite{patel2025tartanground}. We evaluate two sequences per environment across three distinct simulated domains: \textit{Downtown}, \textit{Office}, and \textit{Supermarket}. A characteristic challenge of this dataset is its relatively low visual fidelity and large textureless expanses, which alters the standard dynamics of the visual feature backbone compared to photorealistic datasets.

\subsubsection{Campus Object Dataset (CODa)}
To evaluate the scalability and stability of the system over long-horizon, real-world trajectories, we utilize the Campus Object Dataset (CODa) \cite{zhang2024toward}. Collected via a mobile robot navigating a university campus, this dataset features continuous 15 to 30-minute navigations that include seamless transitions between outdoor street environments and indoor building complexes.

\subsubsection{Train Station (Internal)}
Finally, to evaluate performance in structured but open-plan architecture, distinct from CODa's diverse indoor-outdoor transitions, we recorded a custom dataset within a large, high-traffic transit hub. We evaluate two trajectories navigating from train platforms, up staircases, to a main concourse. Data was collected using a custom platform equipped with a ZED X camera (RGB video and odometry) and a Robosense E1R LiDAR. To obtain high-accuracy RGB-D frames, we backproject the sparse LiDAR measurements into the camera frame and upsample them to the full optical resolution. This environment features large contiguous volumes lacking physical dividing walls but containing strict logical hierarchies and transient crowds.

\subsection{Implementation Details}
\label{subsec:implementation}

Our system is implemented as a standalone Python framework that interfaces with the Hydra spatial perception engine \cite{hughes2022hydra} via ROS 2. 

\subsubsection{Physical Backbone Instantiation}
We utilize the Hydra frontend to extract the topological \emph{place} nodes that form our physical backbone. The extraction mechanism is adapted to the scale of the environment. For highly structured indoor scenes (HM3D and the Train Station), we extract places using a 3D Generalized Voronoi Diagram (GVD). Conversely, for large or unbounded environments (CODa and TartanGround), where 3D GVD extraction becomes unstable or overly dense, we utilize Hydra's 2D place extraction algorithm. To ensure a clean topological representation in these expansive settings, we augment this 2D extraction with a graph filtering step to eliminate outlier place clusters.

\subsubsection{Generative Prior Configuration}
The top-down Prior Graph is generated using GPT-5.4 configured with medium thinking, conditioned on a fixed environment definition $\mathcal{E}_0$. For all quantitative benchmarks, the target \emph{region} set ($\mathcal{R}_P$) is fixed to exactly match the ground-truth classes. This constraint isolates our evaluation of segmentation accuracy from the LLM's open-vocabulary generation variance, though the LLM remains responsible for generating all objects, attributes, and edges. To ensure statistical rigor and account for LLM non-determinism, all reported quantitative metrics are averaged across 10 independent Prior Graph generation runs.

\subsubsection{Multi-Scale Feature Kernel}
We utilize the CLIP ViT-H/14 model \cite{radford2021learning,schuhmann2022laionb} as our default visual backbone. The sensitivity of the system to this foundation model capacity (comparing ViT-H, ViT-L, and ViT-B) is explicitly evaluated in our ablation studies (\Cref{subsec:system_sensitivity}). To extract object-level representations, we query GroundingDINO \cite{liu2024grounding} for open-vocabulary bounding box detections, generate dense segmentation masks from these boxes utilizing FastSAM \cite{zhao2023fast}, and subsequently encode the masked regions using the shared CLIP backbone. For place-level integration, the static feature pyramid is generated with a minimum crop resolution of $224 \times 224$ pixels, scaling up linearly for upper layers. The choice in minimum resolution is also defended in \Cref{subsec:system_sensitivity}. To ensure that the feature kernel only integrates meaningful physical volumes, we enforce a strict relevance threshold: candidate crops are discarded if their 3D projection area is less than $2.25\text{m}^2$. Furthermore, while the semantic query scale $r_Q$ is theoretically a local property of individual \emph{place} nodes, we apply a homogeneous radius per scene for consistency. We set $r_Q=0.75$m for the Train Station, HM3D, and CODa environments. For TartanGround, this radius is increased to $r_Q=1.5$m to account for the lack of meaningful texture in the simulation.

\subsubsection{Expert Activation and MRF Parameters}
The active experts in the Log-Linear Pool are dynamically configured based on the physical capabilities of the datasets (the individual contributions of these experts are quantified in \Cref{subsec:ablation_experts}). For all HM3D environments, all experts (Visual, Geometric, Object, and Environment) are active. The Train Station sequence utilizes the visual, geometric, and environment experts. The object expert is omitted here because massive transit hubs consist of broad architectural zones where discrete object instances are sparsely distributed and lack the dense discriminative power found in residential spaces. Finally, the TartanGround and CODa evaluations rely exclusively on the visual and environment experts. In these expansive indoor and outdoor environments, regions are defined primarily by macro-level visual textures rather than localized object clusters. Additionally, the geometric expert is disabled because local structural inclination does not meaningfully delineate semantic boundaries (e.g., distinguishing a street from a plaza) in these settings.

The energy weights for the Markov Random Field are fixed globally across all experiments to demonstrate that the system does not require scene-specific tuning. The pairwise topological regularizer is set to $\beta=2$. The unary expert reliabilities are set to $\lambda_{\text{vis}}=4$ for vision, $\lambda_{\text{obj}}=1$ for objects, $\lambda_{\text{geo}}=1$ for geometry, and $\lambda_{\text{env}}=0.1$ for the global environment prior.

\subsection{Baselines}
\label{subsec:baselines}

We compare our approach against two state-of-the-art hierarchical scene graph frameworks. These baselines are specifically selected to represent the two orthogonal segmentation paradigms discussed previously: pure bottom-up visual clustering and strict geometric heuristics.

\subsubsection{Clio \cite{maggio2024clio}} 
This framework represents the bottom-up visual clustering paradigm. It partitions physical space into regions utilizing an Agglomerative Information Bottleneck algorithm \cite{slonim1999agglomerative}, which compresses the scene by clustering places that exhibit high information-theoretic similarity. Beyond the clustering algorithm, a fundamental operational difference lies in the visual feature extraction pipeline. Clio processes each incoming 2D image to extract a single global CLIP embedding, which is then uniformly projected onto all 3D places visible within the camera frustum. In contrast, our approach utilizes a multi-scale feature pyramid and a spatial integration kernel to query isolated, correctly scaled visual features for each specific place, preventing the dilution of local semantics by the broader global image context. To ensure a rigorous comparison, we configure Clio to operate with the same Hydra \cite{hughes2022hydra} \emph{place} extraction parameters and utilize the same CLIP ViT-H/14 visual encoder as our method. This configuration isolates the core algorithmic contributions, guaranteeing that any performance variance is strictly attributable to our prior-guided MAP formulation and scale-aware feature pyramid versus their information bottleneck clustering and global feature projection.

\subsubsection{HOV-SG \cite{werby2024hierarchical}} 
This framework represents the geometric heuristic paradigm. It defines regions through structural boundaries, utilizing a watershed algorithm over a 2D Euclidean Distance Field. This method operates on a disjoint pipeline: it first solves for the geometric boundaries based on the assumption that regions are strictly separated by physical constrictions (e.g., doorways and narrow corridors), and subsequently assigns semantic labels to these fixed regions by aggregating visual features from nearby camera frames. We configure HOV-SG to utilize the same CLIP ViT-H/14 backbone for this semantic labeling phase. Because its core segmentation algorithm strictly requires physical constrictions to function, it is fundamentally incapable of segmenting unbounded outdoor environments or open-plan architecture. Consequently, we restrict our evaluation of HOV-SG exclusively to the constrained indoor residential scenes of the HM3D dataset.

\subsection{Evaluation Metrics}
\label{subsec:metrics}

To quantify segmentation performance, we adopt the volumetric metrics introduced by \citet{hughes2022hydra}, evaluating the spatial overlap between the set of estimated regions $\mathcal{R}_I$ and ground-truth regions $\mathcal{R}_{\text{GT}}$ based on 3D voxel occupancy calculated based on the combined volume of its constituent places. 

We evaluate performance across two distinct axes: geometric fidelity ($G$) and semantic accuracy ($S$). To consolidate this formulation, let the effective intersection volume between an estimated region $r_I$ and a ground-truth region $r_{\text{GT}}$ be defined as $I_X(r_I, r_{\text{GT}})$. This intersection volume toggles between pure structural overlap and strictly label-constrained overlap depending on the evaluation axis $X$:
\begin{equation}
    I_X(r_I, r_{\text{GT}}) = 
    \begin{cases} 
      |r_I \cap r_{\text{GT}}| & \text{if } X = G \\
      |r_I \cap r_{\text{GT}}| \cdot \mathds{1}[L(r_I) = L(r_{\text{GT}})] & \text{if } X = S
    \end{cases}
\end{equation}
where $|r|$ denotes the total voxel volume of region $r$, and $\mathds{1}[\cdot]$ is an indicator function that strictly zeroes out the intersection volume if the predicted semantic label does not match the ground truth.

Using this unified intersection term, we compute instance-level volumetric Precision and Recall as:
\begin{equation}
    \text{Precision}_X = \frac{1}{|\mathcal{R}_I|} \sum_{r_I \in \mathcal{R}_I} \max_{r_{\text{GT}} \in \mathcal{R}_{\text{GT}}} \left( \frac{I_X(r_I, r_{\text{GT}})}{|r_I|} \right)
\end{equation}
\begin{equation}
    \text{Recall}_X = \frac{1}{|\mathcal{R}_{\text{GT}}|} \sum_{r_{\text{GT}} \in \mathcal{R}_{\text{GT}}} \max_{r_I \in \mathcal{R}_I} \left( \frac{I_X(r_I, r_{\text{GT}})}{|r_{\text{GT}}|} \right)
\end{equation}
We report the F1 score ($F1_X$) as the harmonic mean of these respective values.
We further report a mean IoU over a Hungarian one-to-one matching $\mathcal{M}$
between the estimated and ground-truth region sets,
\begin{equation}
    \text{mIoU}_X = \frac{1}{|\mathcal{M}|}
        \sum_{(r_I, r_{\text{GT}}) \in \mathcal{M}}
        \frac{I_X(r_I, r_{\text{GT}})}{|r_I \cup r_{\text{GT}}|} ,
\end{equation}
where the matching $\mathcal{M}$ pairs each region with at most one counterpart to maximize the total IoU. Unlike the
overlap-based precision and recall, this measures how tightly each matched region
aligns with its counterpart, isolating boundary quality and, for the
semantic axis, additionally requiring the matched region to carry the correct
label.

\subsection{Comparison with State-of-the-Art}
\label{subsec:sota_comparison}

\begin{figure*}[t]
    \centering
    \includegraphics[width=\textwidth]{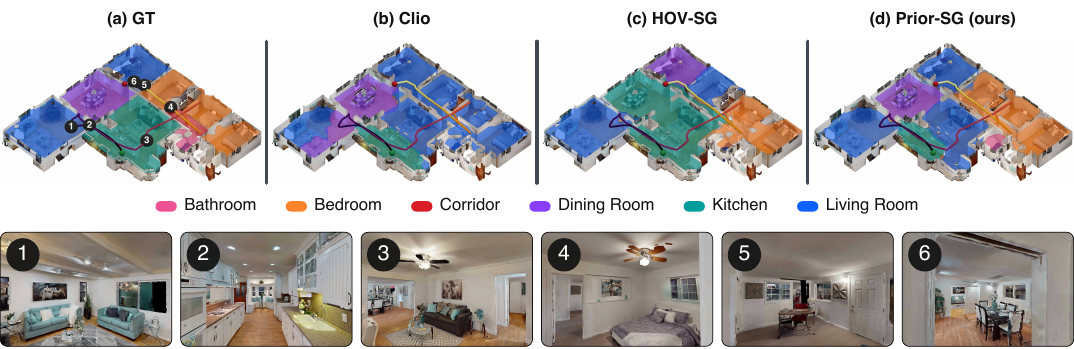}
    \caption{Qualitative comparison of region segmentation on the HM3D-00824 scene. Numbered trajectory markers (1--6) in the (a) ground truth (GT) correspond to the camera views shown in the bottom row. (b) Clio suffers from semantic embedding overlap and semantic smearing, erroneously projecting foreground dining room features (view 6) onto the distant background living room (view 1). (c) HOV-SG fails to segment the unconstricted open-plan kitchen and living area (views 2 and 3) due to a lack of geometric constrictions, and misclassifies regions due to field-of-view overlap in nearby frames, such as assigning dining room features (view 6) to the adjacent living room (view 5). (d) Prior-SG successfully resolves these spatial and semantic ambiguities, utilizing local object context to explicitly separate the open-plan kitchen (view 2) from the living room (view 3), though its topological grouping inherently merges the strictly adjacent bedrooms into a single contiguous region.}
    \label{fig:hmd3_method_comparison_qualitative}
\end{figure*}

\begin{table*}[t]
\centering
\caption{Cross-dataset comparison: Semantic and Geometric Segmentation.}
\label{tab:cross_dataset_combined}
\begin{tabular}{@{}ll|cccc|cccc@{}}
\toprule
 & & \multicolumn{4}{c|}{Semantic Segmentation} & \multicolumn{4}{c}{Geometric Segmentation} \\
\cmidrule(lr){3-6} \cmidrule(l){7-10}
Dataset & Method & P & R & F1 & mIoU & P & R & F1 & mIoU \\
\midrule
\multirow{3}{*}{HM3D (Navigation) \cite{ramakrishnan2021hm3d}} 
 & HOV-SG \cite{werby2024hierarchical} & 53.8{\tiny$\pm$23.1} & 30.5{\tiny$\pm$15.7} & 37.8{\tiny$\pm$18.2} & 46.2{\tiny$\pm$23.5} & 74.4{\tiny$\pm$13.6} & \textbf{90.7{\tiny$\pm$4.4}} & 81.1{\tiny$\pm$9.0} & 65.1{\tiny$\pm$11.0} \\
 & Clio \cite{maggio2024clio} & 46.0{\tiny$\pm$14.9} & 30.6{\tiny$\pm$13.9} & 35.7{\tiny$\pm$11.7} & 23.4{\tiny$\pm$6.6} & 82.3{\tiny$\pm$11.0} & 86.9{\tiny$\pm$3.4} & 84.1{\tiny$\pm$6.4} & 38.1{\tiny$\pm$6.0} \\
 & Prior-SG (ours) & \textbf{71.8{\tiny$\pm$7.9}} & \textbf{65.8{\tiny$\pm$15.9}} & \textbf{68.0{\tiny$\pm$11.2}} & \textbf{61.8{\tiny$\pm$11.6}} & \textbf{85.6{\tiny$\pm$2.5}} & \textbf{90.7{\tiny$\pm$2.7}} & \textbf{88.0{\tiny$\pm$1.9}} & \textbf{65.9{\tiny$\pm$8.0}} \\
\midrule
\multirow{3}{*}{HM3D (Scan) \cite{ramakrishnan2021hm3d}} 
 & HOV-SG \cite{werby2024hierarchical} & 52.1{\tiny$\pm$12.8} & 42.8{\tiny$\pm$12.1} & 46.8{\tiny$\pm$12.0} & 47.8{\tiny$\pm$12.0} & 81.8{\tiny$\pm$8.8} & \textbf{90.1{\tiny$\pm$3.0}} & 85.6{\tiny$\pm$5.6} & \textbf{72.5{\tiny$\pm$8.0}} \\
 & Clio \cite{maggio2024clio} & 35.6{\tiny$\pm$14.5} & 25.9{\tiny$\pm$12.4} & 28.6{\tiny$\pm$11.4} & 18.9{\tiny$\pm$8.5} & 76.3{\tiny$\pm$10.8} & 84.3{\tiny$\pm$4.9} & 79.5{\tiny$\pm$5.7} & 38.1{\tiny$\pm$12.5} \\
 & Prior-SG (ours) & \textbf{74.5{\tiny$\pm$4.6}} & \textbf{65.7{\tiny$\pm$10.8}} & \textbf{69.4{\tiny$\pm$7.4}} & \textbf{62.7{\tiny$\pm$7.8}} & \textbf{86.9{\tiny$\pm$2.6}} & 90.0{\tiny$\pm$3.2} & \textbf{88.4{\tiny$\pm$1.9}} & 69.5{\tiny$\pm$9.1} \\
\midrule
\multirow{2}{*}{TartanGround \cite{patel2025tartanground}} 
 & Clio \cite{maggio2024clio} & 37.1{\tiny$\pm$24.3} & 12.3{\tiny$\pm$7.3} & 13.3{\tiny$\pm$9.6} & 6.0{\tiny$\pm$3.0} & \textbf{92.2{\tiny$\pm$4.4}} & 81.1{\tiny$\pm$17.0} & \textbf{85.2{\tiny$\pm$9.1}} & 24.6{\tiny$\pm$9.1} \\
 & Prior-SG (ours) & \textbf{43.4{\tiny$\pm$12.2}} & \textbf{55.1{\tiny$\pm$26.4}} & \textbf{44.8{\tiny$\pm$15.8}} & \textbf{27.2{\tiny$\pm$12.5}} & 77.9{\tiny$\pm$9.6} & \textbf{90.3{\tiny$\pm$5.6}} & 83.4{\tiny$\pm$7.1} & \textbf{31.0{\tiny$\pm$12.3}} \\
\midrule
\multirow{2}{*}{CODa \cite{zhang2024toward}} 
 & Clio \cite{maggio2024clio} & \textbf{61.3{\tiny$\pm$13.7}} & 30.8{\tiny$\pm$1.8} & 40.5{\tiny$\pm$2.1} & 14.4{\tiny$\pm$2.2} & \textbf{99.4{\tiny$\pm$0.2}} & 86.1{\tiny$\pm$3.0} & \textbf{92.3{\tiny$\pm$1.8}} & 39.2{\tiny$\pm$11.4} \\
 & Prior-SG (ours) & 42.5{\tiny$\pm$12.3} & \textbf{52.9{\tiny$\pm$6.9}} & \textbf{46.8{\tiny$\pm$9.9}} & \textbf{43.8{\tiny$\pm$6.8}} & 78.9{\tiny$\pm$6.6} & \textbf{88.9{\tiny$\pm$3.6}} & 83.4{\tiny$\pm$2.8} & \textbf{50.8{\tiny$\pm$3.7}} \\
\midrule
\multirow{2}{*}{Train Station (Internal)} 
 & Clio \cite{maggio2024clio} & 43.0{\tiny$\pm$27.8} & 51.5{\tiny$\pm$0.8} & 44.2{\tiny$\pm$16.9} & 23.2{\tiny$\pm$27.7} & \textbf{95.8{\tiny$\pm$1.9}} & 80.9{\tiny$\pm$12.7} & 87.5{\tiny$\pm$8.3} & 31.4{\tiny$\pm$16.2} \\
 & Prior-SG (ours) & \textbf{57.3{\tiny$\pm$17.3}} & \textbf{71.5{\tiny$\pm$16.3}} & \textbf{62.4{\tiny$\pm$13.9}} & \textbf{57.9{\tiny$\pm$22.1}} & 90.1{\tiny$\pm$4.1} & \textbf{88.3{\tiny$\pm$6.2}} & \textbf{89.0{\tiny$\pm$3.5}} & \textbf{67.5{\tiny$\pm$12.9}} \\
\bottomrule
\end{tabular}
\end{table*}

We evaluate the performance of Prior-SG against the baseline frameworks across four diverse environments, with multi-dataset quantitative performance summarized in \Cref{tab:cross_dataset_combined} and qualitative performance visualized on HM3D in \Cref{fig:hmd3_method_comparison_qualitative}.



\subsubsection{Quantitative Performance}
As shown in \Cref{tab:cross_dataset_combined}, Prior-SG significantly outperforms all competing methods in semantic segmentation across every evaluated dataset, achieving substantial margins of improvement in both Semantic F1 score ($F1_S$) and Semantic mean Intersection over Union ($\text{mIoU}_S$). Furthermore, while Prior-SG operates on soft probabilistic priors rather than hard structural constraints, it remains highly competitive with dedicated geometric pipelines in classic indoor environments. In the HM3D dataset, our framework matches or exceeds the geometric performance of HOV-SG in natural navigation scenarios and remains tightly competitive during exhaustive scanning ($69.5\%$ vs. $72.5\%$ geometric mIoU), while decisively outperforming it in assigning the correct semantic identities to those geometries.

The division of the HM3D dataset into Navigation and Scan trajectories reveals a critical operational advantage of our framework: robustness to sensor motion. Baseline methods exhibit sensitivity to how the robot moves. For example, HOV-SG experiences a severe drop in semantic F1 performance (falling from $46.8\%$ on scans down to $37.8\%$ on navigation) because its proximity-based labeling heuristic breaks down when the agent does not exhaustively look at every surface. Clio exhibits similar instability across traversal types. In contrast, Prior-SG maintains consistent semantic performance ($69.4\%$ on scans vs. $68.0\%$ on navigation). This stability proves that our prior-guided MAP formulation and scale-aware feature pyramid successfully anchor semantics from partial, natural observations without requiring exhaustive viewpoint coverage.

Finally, the cross-dataset results highlight Prior-SG's strong semantic performance in large-scale environments across both simulation and the real world (TartanGround, CODa, and the Train Station). While Clio achieves competitive geometric F1 scores by broadly grouping physical spaces (e.g., $92.3\%$ in CODa), it struggles significantly with semantic assignment and boundary precision. Prior-SG drastically outperforms Clio in these large, unbounded settings. For instance, our method achieves a semantic mIoU of $43.8\%$ compared to Clio's $14.4\%$ in the real-world CODa dataset, and $57.9\%$ versus $23.2\%$ in the Train Station. This demonstrates that while purely visual clustering can vaguely group relevant geometry, Prior-SG maintains an accurate semantic understanding at scale.

\subsubsection{Qualitative Performance}
The qualitative results in \Cref{fig:hmd3_method_comparison_qualitative} highlight the distinct methodological trade-offs of the evaluated frameworks across two primary operational axes: spatial partitioning logic and feature assignment mechanisms.

Regarding spatial partitioning, the baselines struggle with distinct forms of environmental ambiguity. Bottom-up clustering approaches, such as Clio, are susceptible to embedding overlap in CLIP space. Because the visual feature distributions of different domestic regions, such as a living room and a bedroom, exhibit high cosine similarity under a global encoder, the agglomerative information bottleneck algorithm clusters distinct functional areas into a single spatial partition. Conversely, disjoint geometric pipelines, such as HOV-SG, face structural limitations. Because their spatial partition relies strictly on a watershed algorithm over a 2D Euclidean distance field, they are conceptually incapable of distinguishing between adjacent functional zones that lack physical constrictions. Consequently, HOV-SG fails to find the geometric boundary separating the open-plan living area (visible in view 3), as there is no intervening doorway or corridor.

Beyond spatial partitioning, the baseline frameworks exhibit critical vulnerabilities in how they assign visual features to the 3D map. Clio relies on global feature splatting, which introduces semantic smearing. For example, when the robot directly faces the dining room (near view 6) while the distant living room remains visible in the background, the extracted global CLIP feature is heavily dominated by the foreground dining room context. Uniformly projecting this single global embedding onto all visible 3D places within the frustum erroneously contaminates the distant living room places (visible in view 1) with the dining room identity. HOV-SG, meanwhile, suffers from misclassification due to its viewpoint selection heuristic, which assigns semantic labels by aggregating visual features from the $N$ camera frames physically closest to a region's geometric centroid. During natural robotic trajectories, spatial proximity does not guarantee semantic relevance. In \Cref{fig:hmd3_method_comparison_qualitative}, the spatial region corresponding to the living room (visible in view 5) is erroneously assigned features from the nearby camera frames observing the dining room (view 6), leading the system to mistake the living room for a dining room. Similarly, the frames physically closest to the bedroom centroid (view 4) capture views looking outward into the adjacent living room, causing HOV-SG to misclassify the bedroom as a living room.

In contrast, Prior-SG directly addresses both the structural and representational vulnerabilities of the baselines. To resolve feature assignment errors, our multi-scale feature pyramid and strict projection area thresholds ensure that local visual context is captured accurately without semantic smearing or reliance on proximity-based heuristics. To resolve spatial partitioning ambiguities, our method utilizes top-down layout priors to regularize local visual and object expert inputs. This formulation correctly segments the small bathroom and preserves sharp boundaries between semantically distinguished rooms. While the system groups the directly adjacent individual bedrooms into a single, continuous bedroom region, it demonstrates fine-grained functional distinction in open spaces. Driven by the detected presence of couches and a coffee table in the area opposite the kitchen island (view 3), Prior-SG successfully separates this living room zone from the adjacent kitchen (view 2) rather than extending the kitchen label. Although this architectural distinction is not explicitly encoded in the ground-truth annotations, it demonstrates the system's capacity to infer functional boundaries directly from local object contexts without relying on geometric constrictions or global image feature clustering.

\subsection{Ablation Studies: Expert Synergy}
\label{subsec:ablation_experts}

\begin{table}[t]
\centering
\caption{Expert toggle ablation with semantic scores.}
\label{tab:expert_toggle_semantic_geometric}
\resizebox{\columnwidth}{!}{%
\begingroup
\setlength{\tabcolsep}{4pt}
\begin{tabular}{@{}c|cccccc@{}}
\toprule
 & \multicolumn{4}{c}{Experts} & \multicolumn{2}{c}{Scores} \\
\cmidrule(lr){2-5} \cmidrule(lr){6-7}
Datasets & V & E & G & O & F1$_\text{S}$ (\%) & mIoU$_\text{S}$ (\%) \\
\midrule
\multirow{6}{*}{\shortstack[c]{\textbf{HM3D (Navigation)}\cite{ramakrishnan2021hm3d}\\{\scriptsize Apartments (Simulated)}}} & $\checkmark$ &  &  &  & 64.7{\tiny$\pm$12.7} & 57.7{\tiny$\pm$14.6} \\
 & $\checkmark$ & $\checkmark$ &  &  & 67.2{\tiny$\pm$12.2} & 59.4{\tiny$\pm$14.1} \\
 & $\checkmark$ &  & $\checkmark$ &  & 64.4{\tiny$\pm$12.4} & 57.4{\tiny$\pm$14.2} \\
 & $\checkmark$ & $\checkmark$ & $\checkmark$ &  & 67.0{\tiny$\pm$12.0} & 59.2{\tiny$\pm$13.4} \\
 & $\checkmark$ & $\checkmark$ &  & $\checkmark$ & \textbf{68.0{\tiny$\pm$11.2}} & \textbf{61.8{\tiny$\pm$11.6}} \\
 & $\checkmark$ & $\checkmark$ & $\checkmark$ & $\checkmark$ & 67.7{\tiny$\pm$11.1} & 61.0{\tiny$\pm$11.4} \\
\midrule
\multirow{2}{*}{\shortstack[c]{\textbf{TartanGround \cite{patel2025tartanground}}\\{\scriptsize Diverse Scenes (Simulated)}}} & $\checkmark$ &  &  &  & 44.6{\tiny$\pm$16.7} & 26.9{\tiny$\pm$13.2} \\
 & $\checkmark$ & $\checkmark$ &  &  & \textbf{45.1{\tiny$\pm$15.9}} & \textbf{27.9{\tiny$\pm$12.8}} \\
\midrule
\multirow{2}{*}{\shortstack[c]{\textbf{CODa \cite{zhang2024toward}}\\{\scriptsize Large-Scale Indoor/Outdoor (Real)}}} & $\checkmark$ &  &  &  & 39.5{\tiny$\pm$11.5} & 34.4{\tiny$\pm$8.8} \\
 & $\checkmark$ & $\checkmark$ &  &  & \textbf{46.8{\tiny$\pm$9.9}} & \textbf{43.8{\tiny$\pm$6.8}} \\
\midrule
\multirow{4}{*}{\shortstack[c]{\textbf{Train Station}\\{\scriptsize Large-Scale Indoor (Real)}}} & $\checkmark$ &  &  &  & 58.4{\tiny$\pm$10.0} & 51.9{\tiny$\pm$17.8} \\
 & $\checkmark$ & $\checkmark$ &  &  & \textbf{65.1{\tiny$\pm$9.4}} & 50.8{\tiny$\pm$17.2} \\
 & $\checkmark$ &  & $\checkmark$ &  & 54.1{\tiny$\pm$9.8} & 55.7{\tiny$\pm$20.5} \\
 & $\checkmark$ & $\checkmark$ & $\checkmark$ &  & 62.4{\tiny$\pm$13.9} & \textbf{57.9{\tiny$\pm$22.1}} \\
\bottomrule
\end{tabular}%
\endgroup
}
\end{table}

To quantify the contribution of each heterogeneous signal within our formulation, we evaluate system performance while systematically toggling the constituent experts in the log-linear pool. The results of this evaluation are summarized in \Cref{tab:expert_toggle_semantic_geometric}. Because the multi-scale feature pyramid forms the core semantic representation, the visual expert (V) serves as the foundational baseline across all configurations.

\subsubsection{Topological Prior and Object Influence}
The topological prior (E) enforces the logical structural constraints defined by the hierarchical scene graph. Its standalone impact is substantial: augmenting the visual baseline with the topological prior (V+E) consistently improves performance across all datasets, increasing the semantic F1 score in HM3D (from $64.7\%$ to $67.2\%$), TartanGround (from $44.6\%$ to $45.1\%$), CODa (from $39.5\%$ to $46.8\%$), and the Train Station (from $58.4\%$ to $65.1\%$). This consistent gain validates a core proposition of our framework: regularizing raw, open-vocabulary visual embeddings with top-down logical rules successfully prevents physically improbable region classifications and resolves baseline visual aliasing.

Furthermore, the topological prior serves as the mathematical foundation for object-level reasoning. When this prior is disabled, the environment-to-region, region-to-region adjacency, and region-to-object probabilities default to uniform distributions. Consequently, without the topological prior to explicitly define which objects belong in which rooms, the object expert (O) provides no distinct discriminative signal to influence region classification. For this reason, the object expert is strictly evaluated in conjunction with the topological prior. 

The cascading impact of these local object features is evident in the HM3D dataset, where the addition of the object expert to the visual and topological configuration (V+E+O) achieves the highest overall performance, increasing the semantic F1 score to $68.0\%$ and the mIoU to $61.8\%$. In these constrained indoor environments, objects provide strong discriminative signals that resolve ambiguities where raw architectural visual texture is insufficient. Conversely, the TartanGround and CODa datasets consist of large-scale environments where functional zones are defined primarily by macro-level visual textures rather than localized, object-centric contexts. Because these datasets lack relevant detectable objects, the object expert is omitted, and the integration of visual and topological experts (V+E) achieves the highest overall performance.

\subsubsection{Geometric Constraints}
The geometric expert (G) serves primarily as a proof-of-concept for the framework's modular extensibility. It demonstrates how a simple, specialized plugin can be seamlessly integrated into the Log-Linear Pool to evaluate specific structural rules (e.g., elevation changes or staircases) predicted by the LLM in the Prior Graph. 

Because the implemented heuristic relies on local spatial covariance, its inclusion slightly degrades the raw semantic F1 score across the HM3D navigation trajectories and the Train Station dataset. Natural navigation trajectories do not exhaustively explore the full environments, meaning the underlying topological place graph contains abrupt observational boundaries. These boundary effects inherently introduce noise when computing the geometric descriptors, leading to minor misclassifications at the edges of the explored environment.

However, in the open-plan architecture of the Train Station dataset, the structural value of this plugin becomes evident. While the visual and topological experts (V+E) correctly classify the general vicinity of the functional zones, this loose boundary definition yields a lower mIoU ($50.8\%$). Introducing the geometric expert (V+E+G) explicitly anchors these semantic labels to the physical structural transitions of the staircases, increasing the mIoU to $57.9\%$. This contrast demonstrates that our prior-guided MAP formulation easily accommodates quick, plug-and-play experts to expand system capabilities.

\subsection{Incremental Construction and Open-Plan Robustness}
\label{subsec:incremental_construction}

\begin{figure*}[t]
    \centering
    \includegraphics[width=\textwidth]{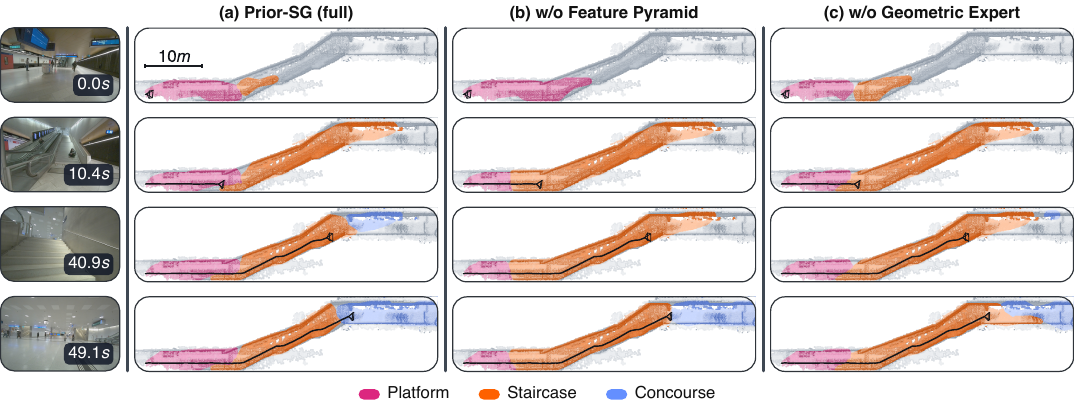}
    \caption{Qualitative online segmentation results along a continuous trajectory in the Train Station 0 scene. The rows represent sequential timesteps ($t=0.0\text{s}$ to $t=49.1\text{s}$), demonstrating the incremental construction of the scene graph. The columns compare the full Prior-SG pipeline against two ablated configurations: one omitting the multi-scale feature pyramid (relying solely on global image features) and one omitting the geometric expert. The full method exhibits robust long-range prediction and crisp boundary delineation, while the ablations demonstrate localized semantic smearing and boundary overestimation.}
    \label{fig:train_station_large}
\end{figure*}
\begin{figure}[t!]
    \centering
    \includegraphics[width=\linewidth]{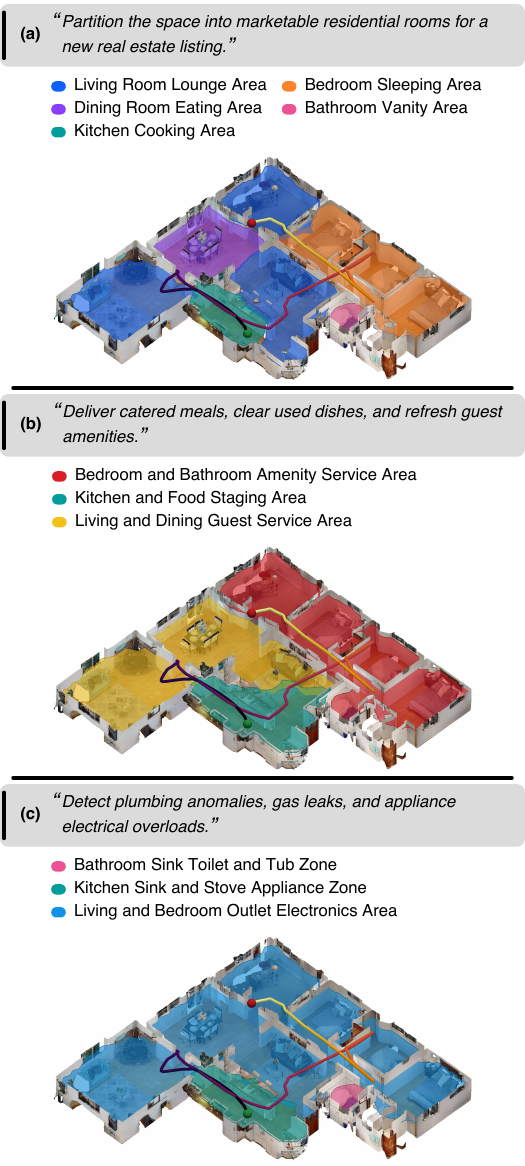}
    \caption{Qualitative demonstration of task-conditioned scene graph generation on the HM3D-00824 navigation scene. Given identical physical geometry and visual observations, the system dynamically restructures its spatial partitioning based on high-level prompts. (a) A real estate listing task recovers standard architectural room definitions (e.g., kitchen, bedrooms, bathrooms), closely mirroring the dataset ground truth in an open-vocabulary manner. (b) A hospitality persona prioritizes service flow, merging the living and dining rooms into a continuous guest area while isolating the kitchen. (c) A facility maintenance persona restructures the map around utility infrastructure, explicitly isolating plumbing and gas zones (kitchen, bathroom) from the remaining electronic living spaces.}
    \label{fig:task_dependence}
\end{figure}

To evaluate the system's capacity to handle large, open-plan architecture where functional zones lack physical dividing walls, we analyze its real-world deployment in a Train Station environment. \Cref{fig:train_station_large} visualizes the online, incremental construction of the hierarchical scene graph across a single continuous trajectory. This chronological breakdown isolates the specific contributions of the feature pyramid and the geometric expert during sequential observation.

\subsubsection{Long-Range Predictive Inference}
The full Prior-SG pipeline demonstrates the ability to accurately segment immediate surroundings while simultaneously predicting distant regions. At $t=0.0\text{s}$, while the robot is positioned on the platform, the system successfully detects and segments a crisp staircase approximately $18\text{m}$ in the distance. As the robot traverses the platform and begins climbing ($t=10.4\text{s}$), the boundary between the platform and the stairs remains highly accurate. Nearing the top of the staircase ($t=40.9\text{s}$), the system predicts the upcoming concourse region prior to fully observing it. Finally, at $t=49.1\text{s}$, the pipeline provides a complete, sharply defined segmentation of the platform, the staircase, and the concourse.

\subsubsection{Impact of the Feature Pyramid}
The necessity of localized feature extraction is demonstrated by ablating the multi-scale feature pyramid and projecting only a single global image feature onto the visible places. At $t=0.0\text{s}$, this variant fails entirely to detect the distant staircase because the global visual context is dominated by the foreground platform. Similarly, at $t=40.9\text{s}$, the system fails to predict the concourse because the global feature is saturated by the immediate staircase geometry. Furthermore, relying on global features introduces severe semantic smearing. As the robot approaches the stairs ($t=10.4\text{s}$), the staircase feature dominates the global context and is erroneously assigned to places in front of the stairs, overestimating its size at the bottom. This localized overestimation persists through the end of the trajectory at $t=49.1\text{s}$.

\subsubsection{Role of the Geometric Expert}
While the visual feature pyramid provides strong semantic cues, the geometric expert serves as a critical structural filter to tighten physical boundaries. When the geometric expert is disabled, the system still successfully detects the staircase at $t=0.0\text{s}$ and $t=10.4\text{s}$ due to the localized visual features, performing significantly better than the global-only baseline. However, it slightly overestimates the footprint of the stairs at the bottom. At $t=40.9\text{s}$, the vision-only system begins to detect a small portion of the upcoming concourse, but it lacks the definitive structural signal required to confidently demarcate the transition. By $t=49.1\text{s}$, this absence of geometric constraints results in boundary bleed, causing the system to noticeably overestimate the total size of the staircase at the top transition into the concourse. The integration of the geometric expert explicitly localizes these boundaries to the physical elevation changes, ensuring crisp transitions in open-plan spaces.

\vspace{-1.0em}
\subsection{Task-Conditioned Flexibility}
\label{subsec:task_condition}

In the preceding quantitative evaluations, the region categories within the prior graph were strictly fixed to align with ground-truth vocabularies, ensuring standardized baseline comparisons. Here, we evaluate the full, unconstrained prior graph generation pipeline. Given a high-level task prompt, the prior graph generator implicitly breaks down the objective to determine a reasonable functional taxonomy, outputting an entirely new, task-specific prior graph complete with novel region categories, adjacency rules, and expected object lists. Subsequently, given the full navigation trajectory, the system constructs the complete instance graph utilizing the identical underlying physical trajectory, place graph, and multi-scale visual features.

As demonstrated in \Cref{fig:task_dependence}, applying three different tasks to the exact same physical trajectory yields distinctly different spatial and semantic partitions. When assigned a standard real estate listing task, the system successfully recovers region definitions that closely mirror the dataset ground truth seen in \Cref{fig:hmd3_method_comparison_qualitative}. However, by shifting to specialized operational personas, such as hospitality or facility maintenance, we observe the framework actively adapting the spatial map, transitioning from human-centric service zones to utility-based infrastructure zones. This example illustrates how \textit{Prior-SG} supports mission-specific spatial reasoning, enabling the robot to dynamically redefine what constitutes a region based directly on the operational requirements of the assigned task, highlighting its potential for zero-shot ontological flexibility.

\vspace{-1.0em}
\subsection{System Sensitivity and Computational Trade-Offs}
\label{subsec:system_sensitivity}
\begin{figure}
    \centering
    \includegraphics[width=\linewidth]{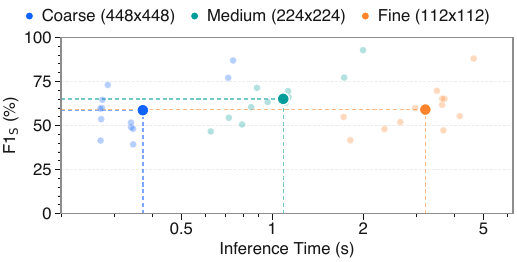}
    \caption{Semantic segmentation F1 score versus per-frame inference time evaluated across different feature pyramid minimum crop resolutions. Each minor dot represents the average performance of 10 trials for a single trajectory, while the large dot denotes the overall mean across all evaluated trajectories for that resolution tier.}
    \label{fig:feature_pyramid_levels}
\end{figure}

To conclude our evaluation, we analyze the computational trade-offs and hyperparameter sensitivity of the feature extraction pipeline. We first evaluate the multi-scale feature pyramid based on its minimum crop resolution (\Cref{fig:feature_pyramid_levels}). The ``Fine'' configuration initiates the pyramid at a 112x112 pixel resolution, yielding a highly granular spatial breakdown. Because the number of required crops scales inversely with the crop area, this fine-grained tiling significantly increases the total number of crops evaluated per frame. This computational burden scales up proportionally with the overall size and expansiveness of the scene. Furthermore, because this resolution falls below the native input requirement of the ViT backbone, the system is forced to artificially upsample each of these numerous crops. This compounding effect introduces interpolative noise and massively inflates the per-frame computation time, resulting in a poor performance trade-off. Conversely, the ``Coarse'' configuration starts at 448x448 pixels, which computes rapidly but misses critical localized details. The ``Medium'' configuration, starting at 224x224 pixels, strikes the optimal balance. It operates at the native ViT resolution, avoiding upsampling latency and bounding the dynamic crop count, while achieving the highest overall F1 score. Beyond these global metrics, this multi-scale extraction strategy computationally justifies itself by unlocking the robust predictive inference capabilities discussed in \Cref{subsec:incremental_construction}.

\begin{figure}
    \centering
    \includegraphics[width=\linewidth]{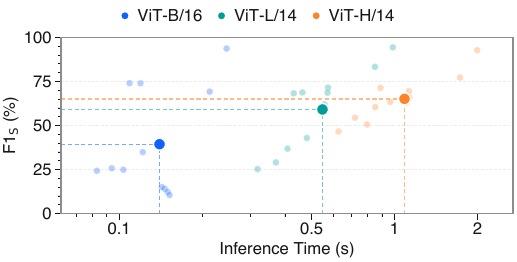}
    \caption{Semantic segmentation F1 score versus per-frame inference time across varying CLIP foundation model capacities. Individual minor dots represent the average of 10 trials for a specific trajectory, and the large dots indicate the global mean performance for each model architecture.}
    \label{fig:clip_backbone}
\end{figure}

We next analyze the sensitivity of the system to the capacity of the underlying CLIP foundation model by plotting segmentation quality against per-frame inference latency (\Cref{fig:clip_backbone}). The lightweight ViT-B/16 model executes rapidly but exhibits highly inconsistent spatial reasoning, rendering it unsuitable for reliable map generation. Both the ViT-L/14 and ViT-H/14 architectures provide the necessary representational stability and achieve comparably high-quality segmentation. While we employ ViT-H/14 as our default visual backbone throughout this work to maintain strict architectural parity with baseline frameworks, the data indicates that ViT-L/14 offers a reasonable compromise between latency and performance for real-time robotic deployment.

\begin{figure}
    \centering
    \includegraphics[width=\linewidth]{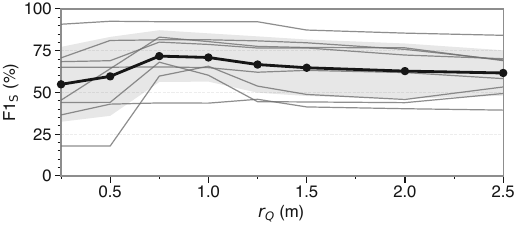}
    \caption{System sensitivity to the spatial query radius ($r_Q$) evaluated on the HM3D navigation dataset. Each faint line corresponds to the performance on one of the eight individual HM3D navigation trajectories. The bold central line indicates the overall mean, with the shaded region representing the $1\sigma$ confidence bounds.}
    \label{fig:r_q_sweep}
\end{figure}

Finally, we evaluate the system's sensitivity to the physical integration parameter, the spatial query radius ($r_Q$), across the HM3D navigation dataset (\Cref{fig:r_q_sweep}). This parameter dictates the spatial volume over which local visual features are aggregated. If the radius is too small ($r_Q < 0.5$m), the system becomes highly susceptible to sparse visual noise and fails to integrate a cohesive regional context. Conversely, if the radius is too large ($r_Q > 1.0$m), the system succumbs to semantic smearing, erroneously dragging visual features across physical boundaries and compromising the ability to cleanly infer unvisited spaces. The performance peak observed at approximately 0.75m validates our empirically chosen parameter, confirming that it optimally balances local detail with regional stability.
\vspace{-1.0em}

%% file: sections/conclusion.tex
\section{Conclusion}
\label{sec:conclusion}
In this work, we introduced \textit{Prior-SG}, a framework that casts hierarchical scene graph generation fundamentally as a probabilistic alignment problem. By moving beyond purely bottom-up reconstruction, we formulated a Maximum A Posteriori (MAP) estimate to align an incrementally built physical map against a task-conditioned structural expectation synthesized by a Large Language Model. To solve this, we developed a multi-scale feature fusion strategy for scale-aware, open-vocabulary perception, alongside a global inference algorithm utilizing graph cuts to fuse local visual, geometric, and object experts with topological priors. We validated this approach across diverse simulated environments and a large, high-traffic real-world transit hub. Our results demonstrated state-of-the-art semantic region segmentation, the ability to robustly predict distant functional boundaries, and provided initial evidence of zero-shot ontological flexibility, allowing the system to restructure spatial partitions based on high-level language tasks.

While these results highlight the potential of prior-guided spatial reasoning, the current framework exhibits specific limitations that define clear directions for future work. Computationally, the graph cut optimization is currently evaluated globally over the observed places at each step. Transitioning this to an incremental inference scheme presents a highly feasible pathway to ensure strict, fixed-compute real-time deployment. Structurally, the inference process is currently limited to a single abstraction layer. To elevate this framework for city-scale missions, additional hierarchical abstraction layers must be introduced to gracefully manage the combinatorial complexity of the alignment. 

Ultimately, future research should focus on evolving the static prior into a dynamic, learning entity. Beyond aligning observations to a fixed expectation, local semantic cues could be used to dynamically update the LLM's internal model. This feedback loop would enable robots to perform true structural extrapolation, allowing them to plan exploration based not just on what is visible, but on a logical certainty of what must lie beyond the horizon.